\documentclass[sigconf,10pt]{acmart}
\graphicspath{{./images/}}
\usepackage[noend,lined,ruled,commentsnumbered,linesnumbered]{algorithm2e}
\usepackage{tabularx}
\usepackage{subcaption}
\usepackage{enumitem}
\usepackage{mdwlist}
\usepackage{caption,subcaption}
\usepackage{mathtools}
\usepackage[noend,lined,ruled,commentsnumbered,linesnumbered]{algorithm2e}
\usepackage{fixltx2e}
\usepackage{caption}                   
\usepackage{booktabs}
\usepackage{multirow}
\usepackage{lipsum}  
\usepackage{graphicx}
\usepackage{xspace}
\usepackage{cuted}
\usepackage{natbib}
\setcitestyle{numbers}
\usepackage{url}
\usepackage{xargs}  
\usepackage{xcolor}  
\usepackage{tikz}

\newcommand*\circled[1]{\tikz[baseline=(char.base)]{
            \node[scale=1.0,shape=circle,draw,inner sep=3.5pt] (char) {};
            \node[scale=1.0] (char) {\textcolor{black}{\footnotesize{\textbf{#1}}}}}}
\newcommand*\blackcircled[1]{\tikz[baseline=(char.base)]{
            \node[scale=1.0,shape=circle,draw, fill,inner sep=3pt] (char) {};
            \node[scale=1.0] (char) {\textcolor{white}{\footnotesize{\textbf{#1}}}}}}

\usepackage{siunitx}
\usepackage{bigdelim}
\usepackage{mathtools}

\newcounter{glossy_enum}
\makecompactlist{list*}{list}

\setenumerate{topsep=0.2em,leftmargin=2.4em,itemindent=-0.0em}

\newcommand{\fakepar}[1]{\vspace{.5mm}\noindent\textbf{#1.}}
\newcommand\figref[1]{Fig.\,\ref{#1}}
\newcommand\secref[1]{Sec.\,\ref{#1}}

\newcommand\beavis{\textsc{BEAVIS}\xspace}

\let\oldabs\abs
\def\abs{\@ifstar{\oldabs}{\oldabs*}}
\let\oldnorm\norm
\def\norm{\@ifstar{\oldnorm}{\oldnorm*}}

\ccsdesc[300]{Hardware~Sensor devices and platforms}


\copyrightyear{2023}
\acmYear{2023}
\setcopyright{rightsretained}
\acmConference[ACM MobiCom '23]{The 29th Annual International Conference on Mobile Computing and Networking}{October 2--6, 2023}{Madrid, Spain}
\acmBooktitle{The 29th Annual International Conference on Mobile Computing and Networking (ACM MobiCom '23), October 2--6, 2023, Madrid, Spain}\acmDOI{10.1145/3570361.3592498}
\acmISBN{978-1-4503-9990-6/23/10}

\begin{document}
\title{\beavis: Balloon Enabled Aerial Vehicle for IoT and Sensing}
\author{Suryansh Sharma$^{\nabla*}$, Ashutosh Simha$^\nabla$, R. Venkatesha Prasad$^\nabla$, Shubham Deshmukh$^\S$, Kavin B. Saravanan$^\S$, Ravi Ramesh$^\S$, Luca Mottola$^{\#}$}
\affiliation{%
  \institution{$^\nabla$Networked Systems, TU Delft, $^\S$Aerodynamics, TU Delft, $^{\#}$Politecnico di Milano and RI.SE}
  \country{}
}

\renewcommand{\shortauthors}{Sharma et al.}
 
\begin{abstract}
UAVs are becoming versatile and valuable platforms for various applications. However, the main limitation is their flying time. We present \beavis, a novel aerial robotic platform striking an unparalleled trade-off between the maneuverability of  drones and the long-lasting capacity of blimps. \beavis scores highly in applications where drones enjoy unconstrained mobility yet suffer from limited lifetime. A nonlinear flight controller exploiting novel, unexplored, aerodynamic phenomena to regulate the ambient pressure and enable all translational and yaw degrees of freedom is proposed without direct actuation in the vertical direction. \beavis has built-in rotor fault detection and tolerance. We explain the design and the necessary background in detail. We verify the dynamics of \beavis and  demonstrate its distinct advantages, such as agility, over existing platforms including the degrees of freedom akin to a drone with 11.36$\times$ increased lifetime. We exemplify the potential of \beavis to become an invaluable platform for many applications. 

\end{abstract}
\maketitle
\vspace{-8pt}
\section{Introduction}
\label{intro} 
Aerial robots represent a new breed of mobile computing platform~\cite{floreano2015science} enabling applications such as aerial mapping and sensing~\cite{UAV-LIDAR-mapping,UAV-3D-mapping, UAV-wifi-sensing,UAV-metrology,UAV-airdrop-remote-sensors} search and rescue~\cite{UAV-search-rescue-2,UAV-search-rescue}, film-making~\cite{UAV-SAR,UAV-human-motion-capture,UAV-3D-images,UAV-radar}, package delivery~\cite{UAV-package-delivery-2} and networking \cite{UAV-LTE}. 
In most cases, their efficient operation in the target scenarios is determined by the combination of two factors: \emph{maneuverability} and \emph{lifetime}. Thus, increasing the lifetime of these vehicles without compromising on maneuverability is a holy grail. 
\begin{figure}[t]
    \centering
    \begin{subfigure}[t]{0.45\linewidth}
        \centering
        \includegraphics[height=\linewidth, width=\linewidth]{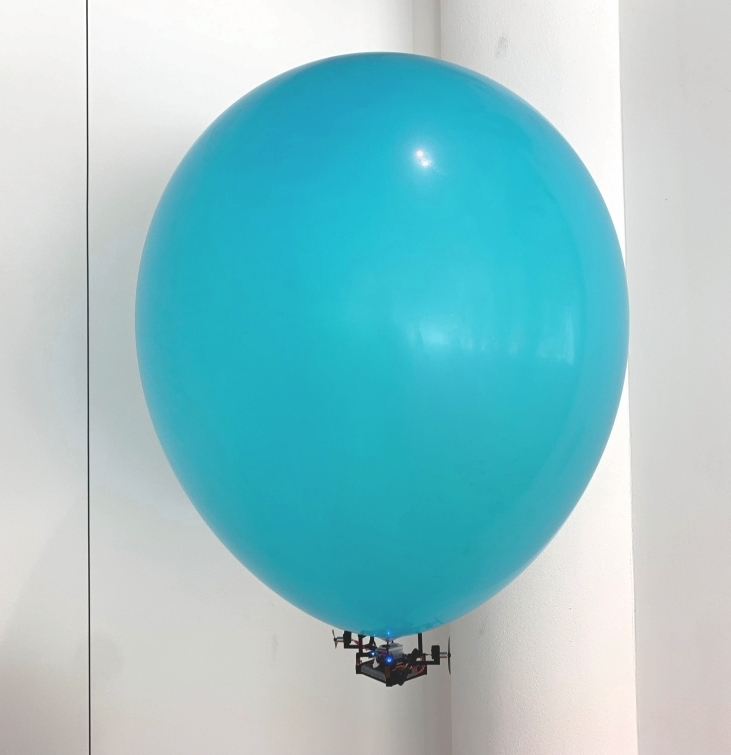}
        \caption{}
    \end{subfigure}
    \begin{subfigure}[t]{0.45\linewidth}
        \centering
        \includegraphics[height=\linewidth, width=\linewidth]{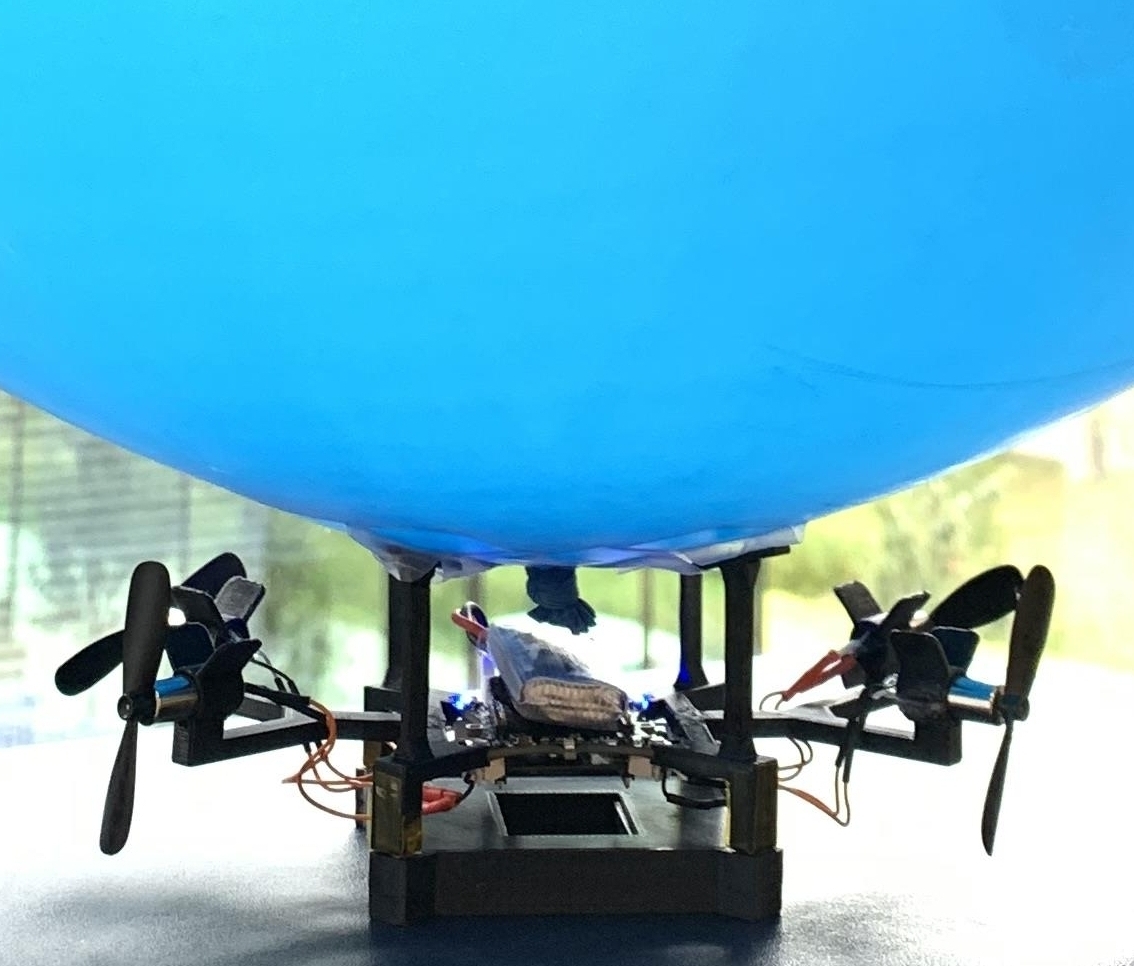}
        \caption{}
    \end{subfigure}
    \vspace{-5pt}
    \caption{BEAVIS prototype: (a)~while flying (b)~rotor orientation close up (video~\cite{video-link}).}
    \label{fig:overview}
\vspace{-20pt}    
\end{figure}

\fakepar{State of affairs} 
\textit{Multi-rotor drones} arguably represent the most widespread aerial robotic platform. Ever-decreasing costs, combined with extreme agility -- both indoors and outdoors -- make them a very useful and important tool to bring sensing and networking where no other platform could reach, for example, in disaster zones, remote and icy mountains, and thick forests~\cite{bregu16reactive}. 
Besides regulatory and safety issues~\cite{afanasov2019flyzone}, multi-rotor drones are also plagued by one major limitation, i.e.,~\emph{the operational time}. Because of the physical dynamics of multi-rotor drones, their motors must produce $\approx$50\% more thrust than the weight of the drone, which is independent of the physical movement in space, if any.
Lifetime is dependent on the battery capacity and thus batteries, being heavy, represent a major weight factor; increasing battery capacity in turn requires higher thrust and high-thrust motors, adding more weight. Thus, the multi-rotor drones are inherently limited by how long they can operate -- reaching hardly a few tens of minutes at best~\cite{afanasov2019flyzone}. 
There are alternatives to multi-rotor drones, yet with limited applicability. \textit{Tethered multi-rotor drones}~\cite{tethered-UAV-6G} obtain energy from a static infrastructure; the necessary electrical connections limit their maneuverability and range.
\textit{Fixed-wing drones}, much like regular aircraft, are extremely difficult to control in constrained spaces, for example, due to their inability to retain a fixed position over time and also because of sophisticated take-off or launch requirements.
They are typically employed only in long-range outdoor missions. \textit{Blimps}~\cite{blimp-open-source}, on the other hand, enjoy long operational times at the expense of limited maneuverability, especially in a vertical motion.

\fakepar{\beavis} In a quest to combine the advantages of different aerial robotic platforms, we design \beavis
 -- \textbf{\underline{B}}alloon \textbf{\underline{E}}nabled \textbf{\underline{A}}erial \textbf{\underline{V}}ehicle for \textbf{\underline{I}}oT and \textbf{\underline{S}}ensing  -- a new platform that combines the maneuverability of multi-rotor drones with the long-lasting hovering capacity of blimps. \figref{fig:overview} shows a \beavis prototype. The structure includes a helium-inflated balloon (or any lighter than air gas, similar to a blimp), attached to a quadrotor structure at the bottom. Unlike a regular quadrotor, however, the motors are not aligned to the vertical plane, but to the \emph{horizontal one}, i.e., propeller blades are vertical to the ground. With the balloon, lift becomes effortless, the horizontal motion requires much less energy, and not all rotors need to operate continuously. 

\fakepar{Intuition} 
The horizontally oriented motors enable movement in any horizontal direction using a single motor. However, the thrust generated by all four creates a high-pressure region in between these rotors which, combined with the dynamic buoyancy effect of the balloon, causes \beavis to rise. When only two opposite rotors are switched on, then \beavis gets lowered. Controlling the four propellers on \beavis yields previously unexplored aerodynamic effects,  which enable motion in three dimensions akin to a regular quadrotor.
Nonetheless, in the absence of external forces, retaining a fixed position requires no propeller motion, as \beavis relies on the lift naturally produced by the balloon. A regular quadrotor needs to spin all propellers merely to hover --consuming almost the same energy as when in motion-- whereas, \beavis consumes almost \emph{no} energy. 

This unique feature allows \beavis to greatly extend the operational lifetime compared to regular multi-rotor drones and represents an asset in all applications where the hovering times of a regular multi-rotor drone would dominate, for example, in surveillance and impromptu networking~\cite{UAV-LTE, UAV-airdrop-remote-sensors, UAV-radar, UAV-wifi-sensing, UAV-3D-images}.
We illustrate the structural design and the flight dynamics of \beavis in \secref{sec:system} and \secref{sec:CFD}, respectively. The structural design addresses the trade-off between size and weight without requiring any more than four actuators to limit size, complexity, and costs. 

Since the flight dynamics of \beavis are not known and we are the first to use an unexplored aerodynamic effect for vertical motion, designing the flight control logic is highly challenging that was further complicated by highly nonlinear flight dynamics and additional drag introduced by the balloon. Our investigation of the flight controller for \beavis is data-driven. We illustrate how a data-driven model-free control logic can be used for autonomous flight (\secref{sec:controller}). We also demonstrate how using our new algorithm, we can, not only sense faults in the rotors but also provide an adaptation of our controller to fly the platform \textit{despite} rotor failure.

\fakepar{Contributions and results}
\blackcircled{1} We provide the unique system design showing how \beavis is built and can be customized as dictated by the applications (Sec.~\ref{sec:system-design}).
\blackcircled{2} We provide a detailed characterization and explanation of the novel aerodynamics of \beavis platform (Sec.~\ref{sec:CFD}).
\blackcircled{3} We design a new data-driven flight control algorithm specifically for \beavis with a focus on fault detection and recovery (Sec.~\ref{sec:controller}).
\blackcircled{4} We put \beavis through its paces and provide evidence of its performance along key application-level performance metrics, such as tracking accuracy and response time. We demonstrate this in two different application scenarios (Sec.~\ref{sec:app-eval}).
\blackcircled{5} We complement this evaluation with the evaluation of system performance using metrics -- including translation and rotational speed, and lifetime -- to provide grounds for a more general understanding of \beavis performance (Sec.~\ref{sec:system-metrics}). 

Our results indicate that we can achieve up to 1136.3$\%$ increase in flight time with complete planar, vertical, and yaw control possible (video:\cite{video-link}). We can fly up to 2.45\,m/s in the \texttt{X-Y} plane and can spin at \ang{346}/s rotational speed. Our algorithm can detect rotor faults within 5.5\,s for 90\% of the time and successfully fly despite them. We also demonstrate the application of people tracking with an 80\% success rate and that of rooftop thermal imaging where we can adapt to varying heights of the roofs within 2.1 seconds (\secref{sec:eval}).
\vspace{-5pt}
\vspace{-2pt}
\section{Related work}
\label{sec:related}
The popularity of UAVs is often diminished by their severely limited flight time which is critical for wireless sensing applications. In recent literature, there have been multiple workarounds proposed for this constraint of UAV platforms. 

\begin{figure}[!tb]
    \centering
    \includegraphics[width=\linewidth]{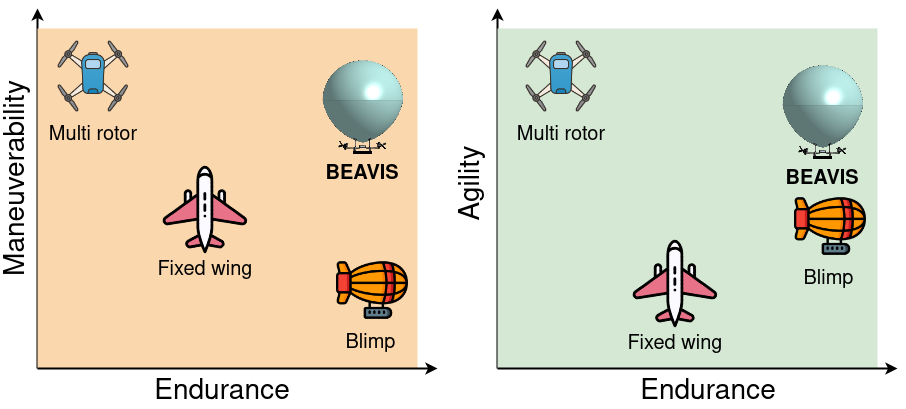}
    \caption{BEAVIS platform comparison for endurance/maneuverability and endurance/agility.}
    \label{fig:trade-off}
\vspace{-15pt}
\end{figure}

\fakepar{Pushing the limits of UAVs} Research efforts exist which optimize UAV flight time and data collection strategy. Gong et al.~\cite{UAV-flight-time-minimization-sensors} minimize a UAV's total flight time (while cruising or hovering) to collect data from sensors on the ground in lesser time. 
Modulating the speed of the UAV to fairly collect data in an Intelligent Transportation System (ITS) has also been done~\cite{UAV-data-collection}. Lee et al.~\cite{UAV-free-space-optics} use a fixed-wing UAV equipped with a free space optical communication (FSOC) system to perform trajectory optimization for given data rates. The common thread in these diverse research is that they all push the utilization of the limited UAV battery to its maximum for wireless sensing and communication applications. Despite their efficient use of the battery, they all are bound by the same energy budget which a limited battery provides. An attempt to increase flight time by using a staged battery approach~\cite{Energy-staging} has been proposed where a battery stage is ejected in-flight after its depletion. 
While a promising idea, dropping depleted batteries has practical implications to consider. Other than optimizing the speed, trajectory, or battery usage, some researchers also extend UAV lifetime by proposing wireless battery charging methods~\cite{wireless-charging}. 
There is also work which studies the dynamics of laser-charged UAVs using a low-power laser source to harvest energy~\cite{laser-charging}. These require elaborate ground station setup and thus, are highly expensive and not feasible for remote locations. 

\fakepar{Alternative platforms} Many alternative aerial platforms like tethered drones, high-altitude weather balloons, and blimps (or airships) have also been explored in place of conventional multi-rotor or fixed-wing UAVs. Cellular base stations (BSs) carried by tethered drones were proposed to circumvent the limited energy and flight time of UAVs~\cite{tethered-UAV-6G}, however traditional UAVs need to land for recharging, which limits their performance. This, however, comes at the expense of reduced mobility~\cite{tethered-drones-coverage} as these platforms have a fixed range as well as a maximum altitude of operation. This gives rise to a trade-off to be made between mobility and endurance. It is also worth noting that controlling and stabilizing tethered single and multi-UAV systems is non trivial~\cite{control-tethered-uav} due to the dynamics introduced by the tether.

A close derivative of High Altitude Platform Stations is the class of aerial vehicles called blimps (or airships) which are lighter-than-air high endurance platforms with 1-6 propellers that enable directed motion. Gonz\'alez-Álvarez et al.~\cite{blimp-open-source} present the design of a lightweight, low-cost, open-source airship along with an accurate altitude control system. 
\textcolor{black}{These and other similar platforms like Blimpduino~\cite{blimpduino-open-source} fare very well in terms of endurance, but this comes at the cost of limited agility and maneuverability.} 
Further miniature autonomous blimps experience swing oscillations due to their often under-actuated design and aerodynamic shape. These can be reduced to some extent with intelligent control systems design~\cite{blimp-swing-Gtech} but their range of motion remains small as a result of their structural design. In particular, vertical and omnidirectional motion is a challenge. \textcolor{black}{In designs like X4-blimp~\cite{blimp-four-envelop} and H-aero~\cite{blimp-H-aero} vertical (rise and descent) motion was made a little easier by the use of a four partial envelope design of the balloon.}
To make such platforms more agile and maneuverable, certain hybrid balloon-drone designs have been tried in literature. 
Skye~\cite{spherical-onmi-blimp} is one such platform that has a spherical blimp design with four tetrahedrally-arranged motors. It is capable of omnidirectional motion with control in six degrees of freedom similar to a drone. This was substantially better than any traditional blimp design could achieve. The key features of its design are safety, maneuverability, and agility. Its extremely complex and large mechanical structure and balloon design (2.7\,m wide) however, makes it suitable only for its intended application of entertainment robotics. It makes it inaccessible to other domains where serious engineering effort with complex control is required. Another innovative hybrid design is ZeRONE~\cite{indoor-blimp-CHI}, an indoor helium-drone-hybrid drone that uses the wind generated by vibrating piezo elements for propulsion instead of using rotors. The biggest takeaway is that this drone design would be safer compared to the alternatives but this comes at the cost of reduced speed, little to no resistance to wind flow, and a very short control-able flight time. 
Our platform \beavis in contrast achieves a trade-off between endurance, agility, and maneuverability (Fig.~\ref{fig:trade-off}). \textcolor{black}{Maneuverability is the degrees of freedom of an UAV that allow it to maneuver; agility is the minimum space required when flying at the minimum allowed speed to accomplish a maneuver~\cite{nano-blimp-lifetime}.}
\vspace{-5pt}
\section{Platform Design}
\label{sec:system}
\begin{figure}[tb!]
    \centering
    \includegraphics[width=0.8\linewidth]{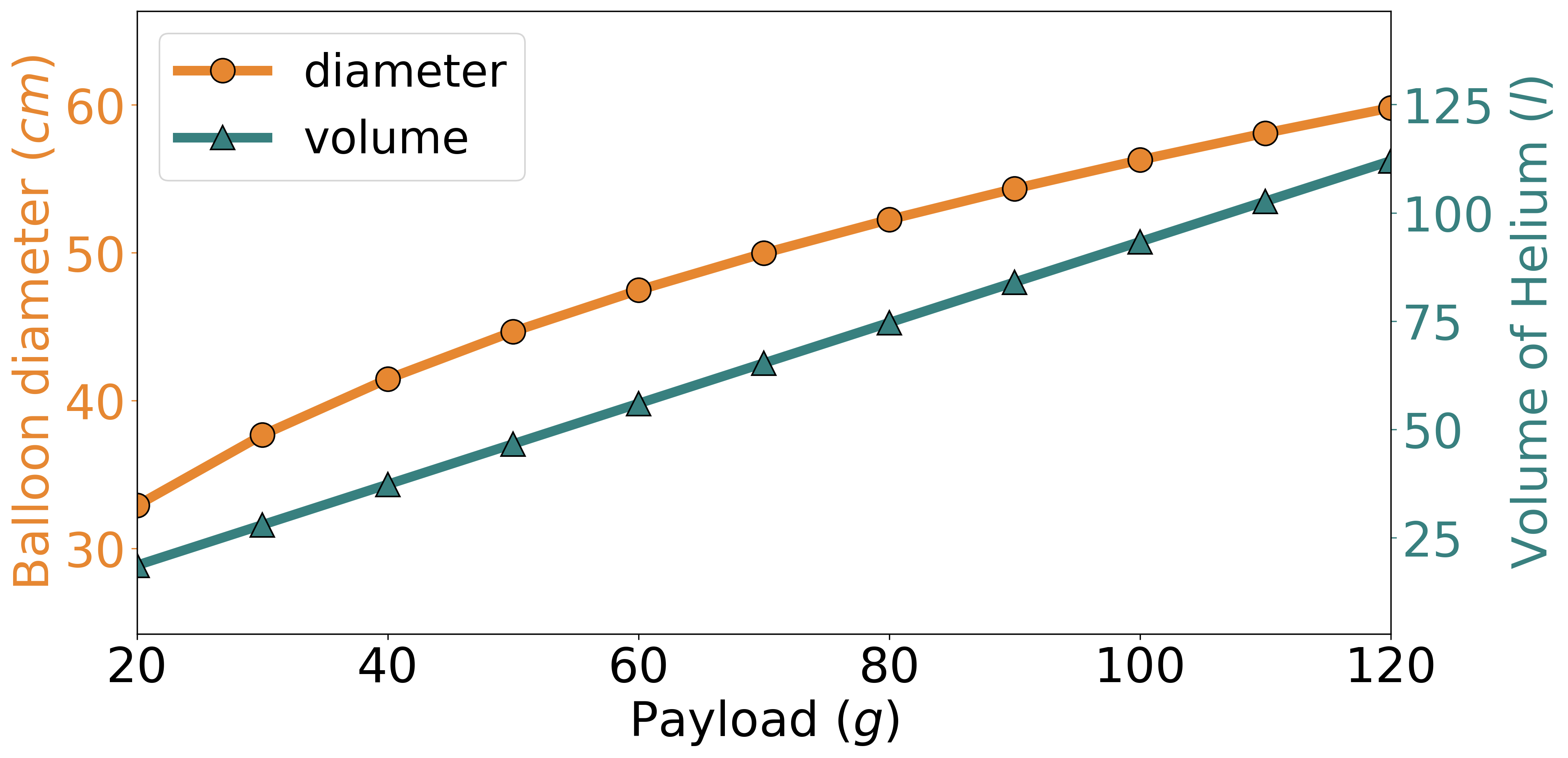}
    \caption{Scaling of the volume of helium required and the size of the balloon with increasing payload.}
    \label{fig:helium-size}
    \vspace{-10pt}
\end{figure}
The goal of \beavis is to offer a platform that can provide both maneuverability (similar to a drone) and large flight time by leveraging the lift produced by a helium-filled balloon. This combined with its focus on sensing and various IoT applications -- both indoors and outdoors -- requires maximizing the carried payload. Thus the addition of any redundant actuators to add agility will result in extra weight, higher complexity, and an increase in the balloon size. 

\textcolor{black}{Fig.~\ref{fig:helium-size} illustrates the balance between payload-to-lift and helium required to achieve neutral buoyancy at 10\,m from the ground. We can see the balloon’s diameter scaling with increasing payload for a regular latex balloon. The aim is to keep \beavis as light as possible to maximize the carried payload for a given balloon size. In this work, we made a miniature indoor version of \beavis to demonstrate its capability in a controlled environment. We are limited in conducting experiments outdoors as a consequence of our location's proximity to a working airport. However, using Fig.~\ref{fig:helium-size}, one can scale up the design based on the requirements.}

The design goals of \beavis can be summarized as follows:
\fakepar{\blackcircled{1} Endurance} The platform must include an inflated helium balloon which will provide lift and enable hovering at a stable altitude with no additional used power. This removes the requirement for powering any actuator for maintaining its height at an equilibrium.

\fakepar{\blackcircled{2} Maneuverability} The platform must have the freedom to move in all directions(omnidirectional) in the horizontal plane when hovering at a stable height. Further, it must be possible to correct its heading and altitude whenever needed. There must be a provision to move in all degrees of freedom as a traditional UAV.

\fakepar{\blackcircled{3} Optimized size/weight} The platform must weigh as less as possible to maximize payload that can be carried for a given balloon size.

\fakepar{\blackcircled{4} Resilience} The platform must function despite one or more rotor failures with the capacity to sense the failure and fly despite it. 
\textcolor{black}{This adds safety and robustness to the system by ensuring that the vehicle does not crash, can finish its mission, and be recovered if a rotor fails.}

\fakepar{\blackcircled{5} Simple design} The platform should be accessible, easily replicated, and beneficial to the sensing and wireless community. This means no complex actuator or mechanical design.

\vspace{-12pt}
\subsection{System Design}
\label{sec:system-design}
\begin{figure}[!tb]
    \includegraphics[width=0.75\linewidth]{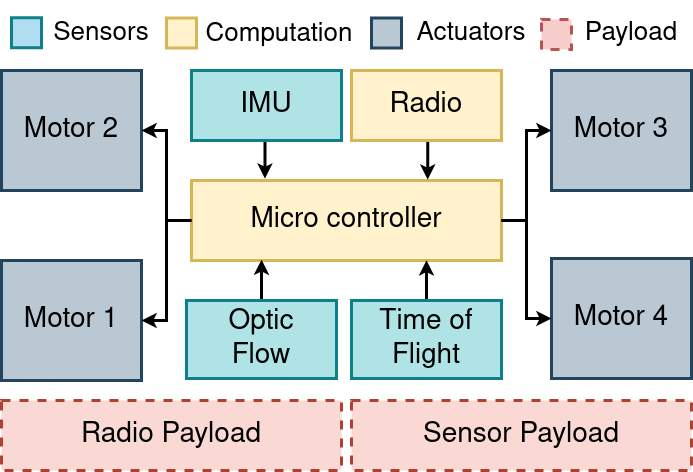}
    \caption{BEAVIS system components with payload.}
    \label{fig:system}
\vspace{-10pt}
\end{figure}

We consider the fundamental design choices and constraints mentioned above while designing \beavis. It consists of three basic system blocks: sensors, computation units, and actuators as shown in Fig.~\ref{fig:system}. Apart from building the \beavis platform, we have also mounted some payloads to demonstrate applications.  
The platform uses three sensors to get an estimate of its attitude, altitude, and heading. It uses an Inertial Measurement Unit (IMU) to infer its accelerations as well as angular velocities. We used a BMI088 which is a 6-axis inertial sensor with a highly accurate 16-bit digital, triaxial accelerometer and a 16-bit digital, triaxial gyroscope. 
To accurately measure its altitude we used a downward-facing Time of Flight (ToF) sensor mounted on the bottom of the platform which can sense the height at which the platform is hovering from the object below (usually the ground). Specifically, we used VL53L1x which is an infrared laser-ranging sensor with accurate ranging up to 4\,m, and fast ranging frequency of up to 50\,Hz. For estimating the translational velocity of the platform a PMW3901 optic flow sensor was used. The PMW3901 sensor uses a low-resolution in-built camera to detect the motion of surfaces to sense \texttt{x-y} motion.

The STM32F405 microcontroller is the main processing unit for controlling the drone autonomously and giving the desired motor commands using the sensors. We programmed the microcontroller for the realtime execution of our algorithm for autonomous flight. To communicate with the platform, for logging as well as for over-the-air firmware flashing, a low-power, long-range radio was used. We used a 2.4\,GHz nRF51822 SoC for this purpose. The radio was communicating with a ground station connected to the PC.
\begin{figure}[!tb]
    \centering
    \begin{subfigure}{0.48\linewidth}
        \centering
        \includegraphics[width=\linewidth]{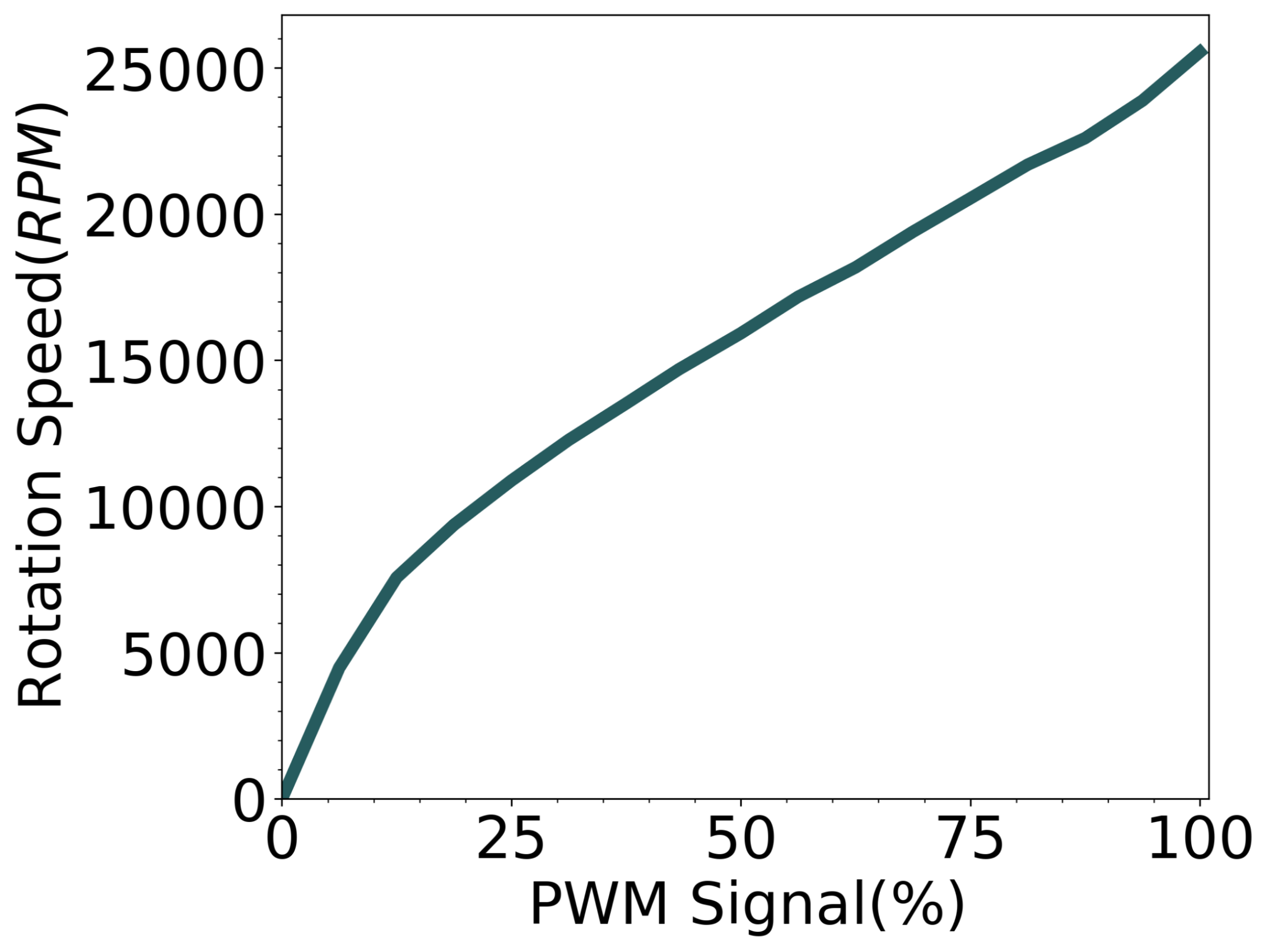}
    \end{subfigure}
    \begin{subfigure}{0.48\linewidth}
        \centering
        \includegraphics[width=\linewidth]{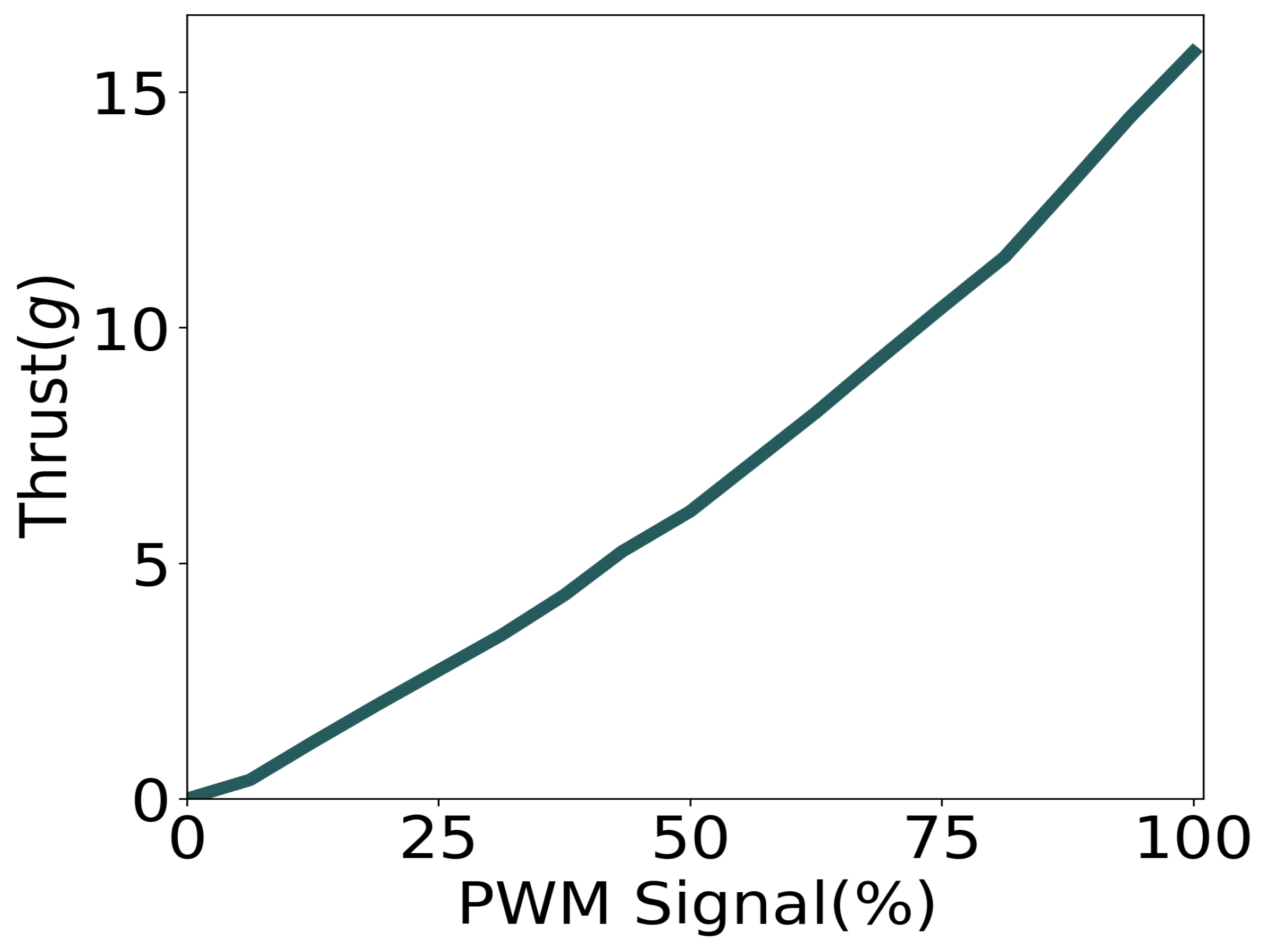}
    \end{subfigure}
    \caption{Variation of motor rotation speed and thrust as a function of motor PWM.}
    \label{fig:motor-character}
\vspace{-10pt}
\end{figure}

The main actuators for achieving all the motion of \beavis are four core-less DC motors equipped with propellers. In our prototype we used 7 x 16\,mm DC motors with a maximum measured rotation speed of 26000\,RPM and a maximum thrust generated by a single motor is 15.7\,g. This can be seen in Fig.~\ref{fig:motor-character} which shows characterization obtained through experiments. The motors were rated for 1\,A of current, and the whole electronic system was powered by a 300\,mAh 4.2\,V LiPo battery. Two motors spun clockwise and two spun anticlockwise. 

\vspace{-5pt}
\subsection{Structural Design}
\label{sec:structural}
\begin{figure}[!tb]
    \centering
    \begin{subfigure}{0.45\linewidth}
        \centering
        \includegraphics[width=\linewidth]{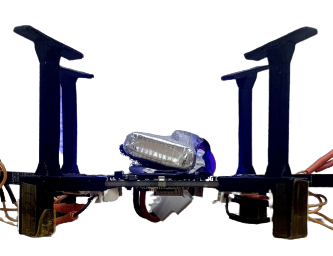}
    \end{subfigure}
    \begin{subfigure}{0.45\linewidth}
        \centering
        \includegraphics[width=\linewidth]{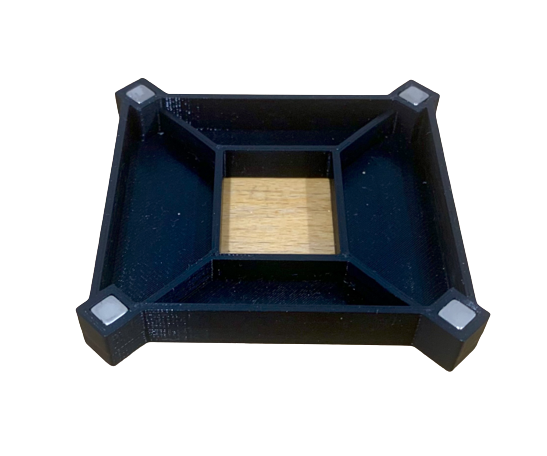}
    \end{subfigure}
    \caption{Balloon mount (left) and detachable ballast (right) to adjust balloon buoyancy.}
    \label{fig:ballast-balloon}
\vspace{-10pt}
\end{figure}

The mechanical mounting structure of \beavis is described here. To generate the lift required to lift the platform (and any attached payload) we use a large latex balloon of size 90\,cm. The balloon is filled with helium based on the total lift-off weight of the platform and payload. This resulted in the platform having neutral buoyancy at a desired predetermined height. Since the process of inflating a helium balloon without the use of a gas pressure gauge is not extremely precise, we designed a detachable ballast to fine-tune the buoyancy of the platform. The ballast shown in Fig.~\ref{fig:ballast-balloon} is attached to the platform via four neodymium N45 magnets and has empty slotted spaces for adjusting the weight of the platform to match the lift generated by the helium in the balloon. The role of the slotted chambers is to adjust if required, the centre of mass of the whole structure to account for any misalignment in weight distribution caused by the addition of payload.
The helium-filled balloon is attached directly above the PCB using four 15\,mm long 3D printed flaps angled at \ang{15} from the horizontal plane. The balloon is mounted at a height of 20\,mm from the PCB leaving this space for interaction of the incoming flows from the four motors which are extremely essential to the design of \beavis. The balloon mount is shown in Fig.~\ref{fig:ballast-balloon}.

The four motors are placed horizontally in a plane with all of the propellers facing outwards. The motors are capable of steering \beavis in any direction in the horizontal plane, unlike a traditional blimp. This means that at any instant \beavis can move in any horizontal direction. This arrangement of motors was done to give an omnidirectional capability to the platform. Further, this arrangement also enables vertical motion, as explained below. 

When motors are spun they pull the air inside towards the center of the PCB. We observe two very interesting natural phenomena depending on the configuration: \circled{1}When all four motors are spun there will be a high-pressure region created in the center between the PCB and the balloon. This causes the whole system to rise, increasing its altitude. \circled{2}When two diagonally opposite motors are activated there will be a low-pressure region created instead which will move the platform down, decreasing its altitude. We exploit this flow interaction from the motors to add a degree of motion to the system without the use of any extra actuators facing in the up-down direction. This results in \beavis being able to translate in a horizontal plane as well as control its altitude. 
\begin{figure}[!tb]
    \centering
    \includegraphics[width=0.8\linewidth]{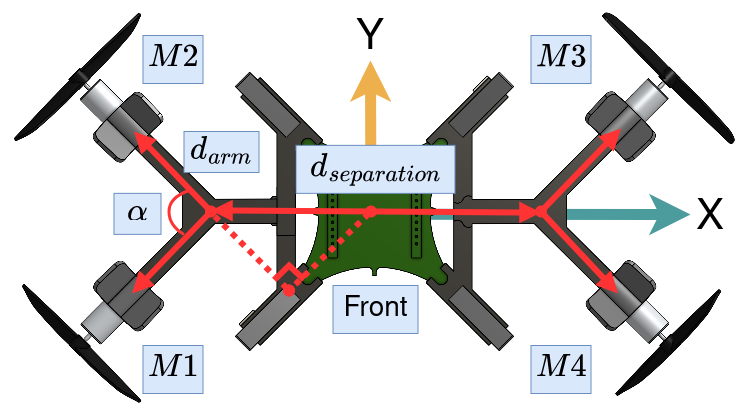}
    \caption{Novel motor arrangement for \beavis to enable horizontal, vertical and yaw control.}
    \label{fig:motor-mount}
\vspace{-5pt}
\end{figure}

\begin{table}[!tb]
\caption{Weights of system components of \beavis as used in our prototype including example payload}
\label{tab:weight-BEAVIS}
\resizebox{\columnwidth}{!}{%
\begin{tabular}{@{}llcl@{}}
\toprule
Part                                        & Overview                                                  & Number                 & Weight(g)         \\ \midrule
\multicolumn{1}{l}{Balloon}               & \multicolumn{1}{l}{90\,cm latex ballooon}                 & \multicolumn{1}{c}{1} & \multicolumn{1}{l}{25.9\,g} \\ \midrule
\multicolumn{1}{l}{Flight controller PCB} & \multicolumn{1}{l}{Computation unit PCB with IMU}        & \multicolumn{1}{c}{1} & \multicolumn{1}{l}{7\,g} \\ \midrule
\multicolumn{1}{l}{Height control PCB}    & \multicolumn{1}{l}{Time of flight and optic flow sensor} & \multicolumn{1}{c}{1} & \multicolumn{1}{l}{2.3\,g} \\ \midrule
\multicolumn{1}{l}{Battery}               & \multicolumn{1}{l}{4.2\,V LiPo 300\,mAh}                    & \multicolumn{1}{c}{1} & \multicolumn{1}{l}{9.7\,g} \\ \midrule
\multicolumn{1}{l}{Motors}                & \multicolumn{1}{l}{DC motors with propellers}            & \multicolumn{1}{c}{4} & \multicolumn{1}{l}{11.9\,g} \\ \midrule
\multicolumn{1}{l}{3D Printed frame}      & \multicolumn{1}{l}{PLA plastic with 15\%infill}          & \multicolumn{1}{c}{1} & \multicolumn{1}{l}{12.0\,g} \\ \midrule
\multicolumn{3}{r}{\textbf{Total platform weight}}                                                                                                        & 68.8\,g  \\ \midrule 
\multicolumn{1}{l}{Payload}               & \multicolumn{1}{l}{Ballast simulating payload}           & \multicolumn{1}{c}{1} & \multicolumn{1}{l}{57.4\,g} \\ \midrule
\multicolumn{3}{r}{\textbf{Total take-off weight}}                                                                                                        & 126.2\,g  \\ \bottomrule
\end{tabular}%
}
\end{table}

\begin{figure}[tb!]
    \centering
    \includegraphics[width=0.69\linewidth]{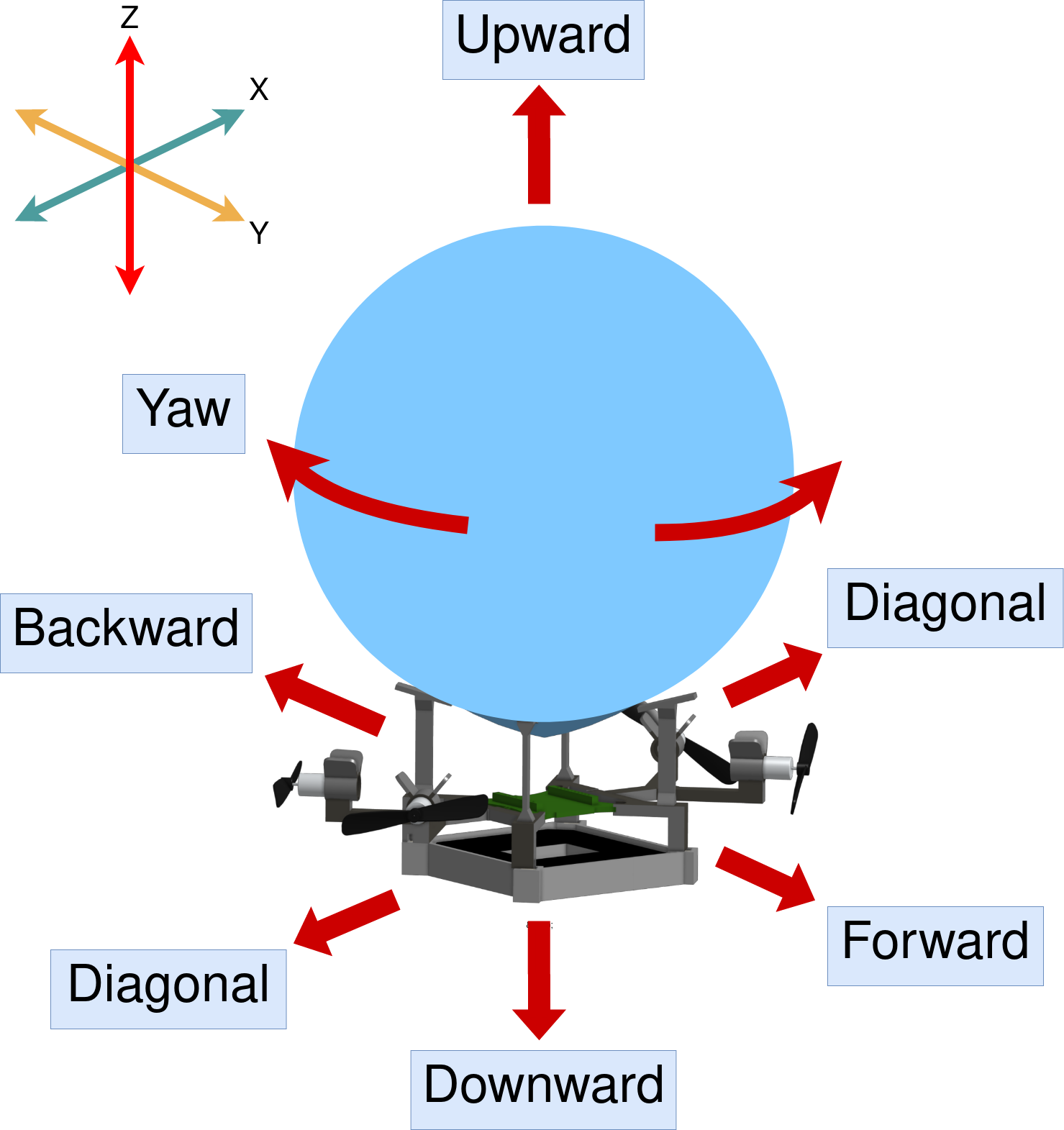}
    \caption{\beavis design with all components and degrees of freedom possible.}
    \label{fig:BEAVIS-CAD-all-DOF}
\vspace{-8pt}
\end{figure}

In addition to moving horizontally and vertically, we design the motor mounts such that we can also control the yaw rotation of the platform. This is particularly useful while using a downward-facing camera or a directional antenna as a payload. In these scenarios, the platform must be able to control (and hold) its heading or yaw angle. We add yaw control by adding a distance $d_{\text{separation}}$ between the two pairs of motors. The motors themselves are mounted at an angle $\alpha$. In Fig.~\ref{fig:motor-mount}, we see the difference in our design compared to the default approach of placing the motors in the shape of an '\texttt{X}'. By separating the motor at a separation of distance $d_{\text{separation}}$, we can calculate the maximum angular moment using,
\begin{equation}
\vspace{-5pt}
M_i = T_i  d_{\text{separation}}  \sin(\alpha/2),   
\end{equation}
where $M_i$ is the moment due to thrust $T_i$ for Motor $i$, $\alpha$ is the angle between two adjacent motor arms, and $d_{\text{separation}}$ is the distance between the motors. While the resultant moment can be increased by increasing $\alpha$, the lateral force component decreases. Therefore, one needs to find an optimal value based on the requirement of turning torque (affects yaw) and lateral force (affects translation). In our prototype, the value of $d_{\text{separation}}$ was set to 80\,mm and $\alpha$ was set to \ang{90}. When combined, the overall design of \beavis enables the movement of the platform as depicted in Fig.~\ref{fig:BEAVIS-CAD-all-DOF}. The components and their weight as well as the possible payload using a helium balloon, inflated to 55\,cm, is given in Table~\ref{tab:weight-BEAVIS}. %
\vspace{-5pt}
\section{Computational Flow Simulation and Analysis}
\label{sec:CFD}
We conduct a computational fluid dynamics (CFD) study to understand and characterize the unique flow interaction that happens in \beavis by visualizing the movement of air fluid around each propeller. We provide a brief note on this to confirm and prove that our design can be generalized.
\begin{figure}[tb!]
    \begin{subfigure}{0.49\linewidth}
        \centering
        \includegraphics[width=\linewidth]{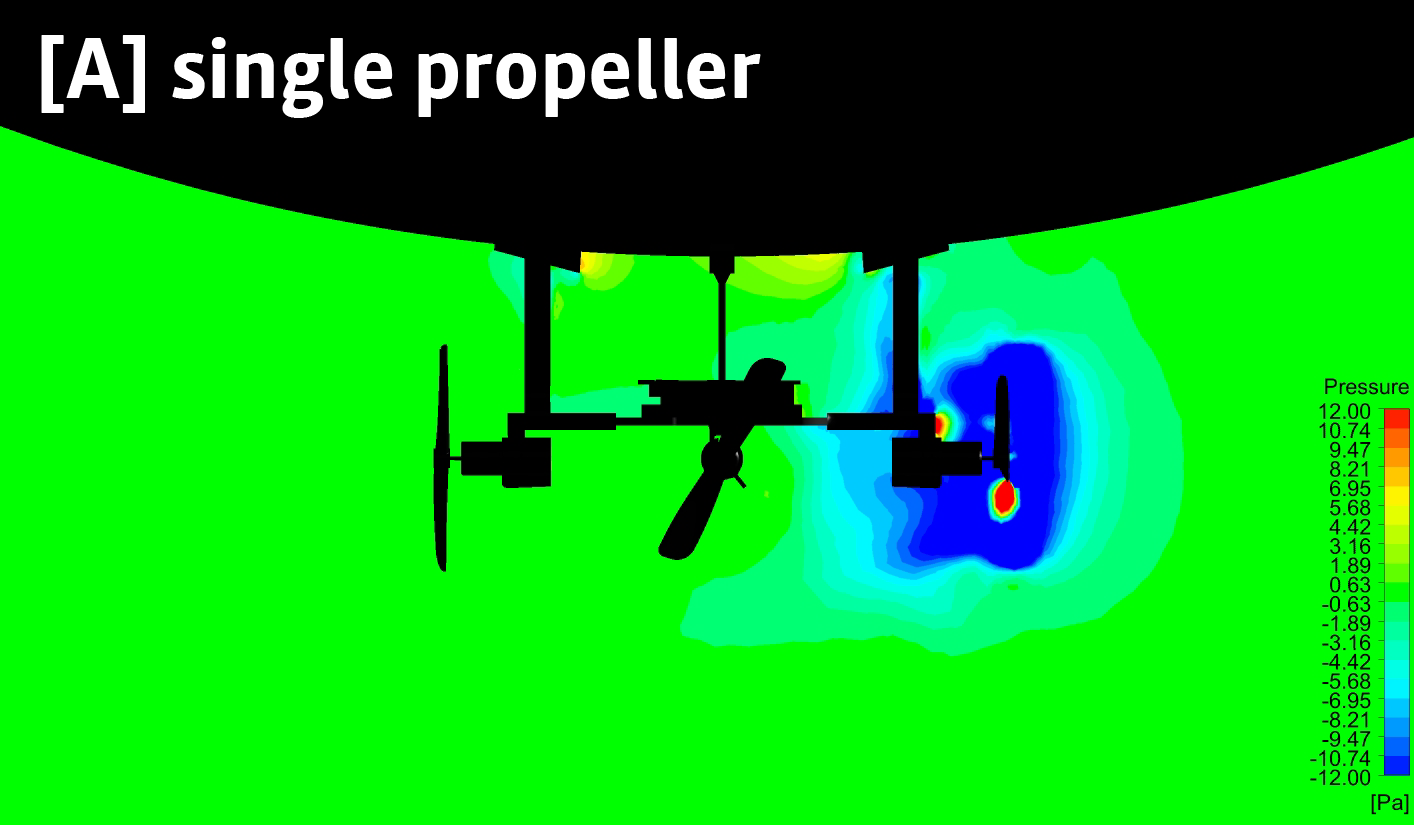}
    \end{subfigure}
    \begin{subfigure}{0.49\linewidth}
        \centering
        \includegraphics[width=\linewidth]{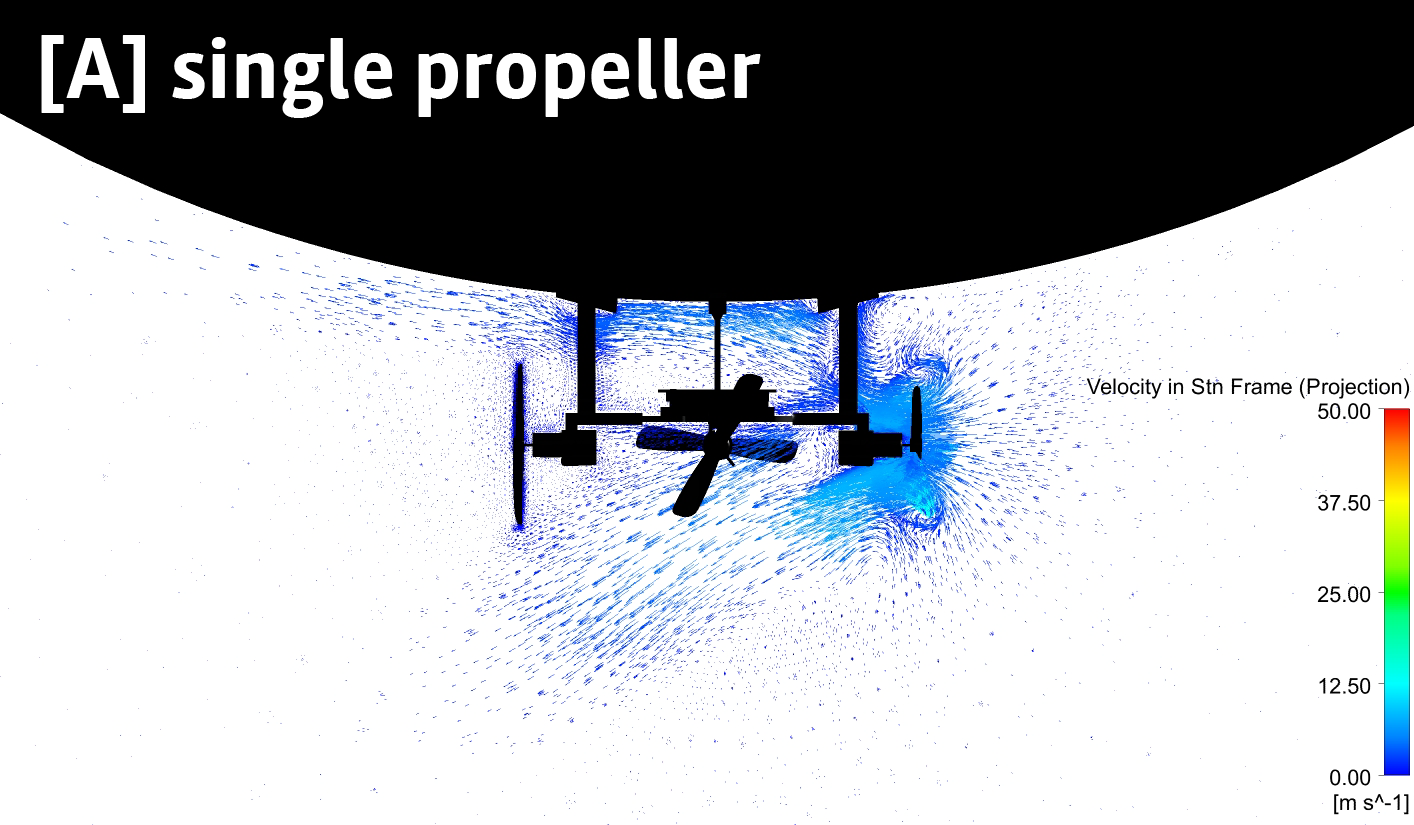}
    \end{subfigure}
    \begin{subfigure}{0.49\linewidth}
        \centering
        \includegraphics[width=\linewidth]{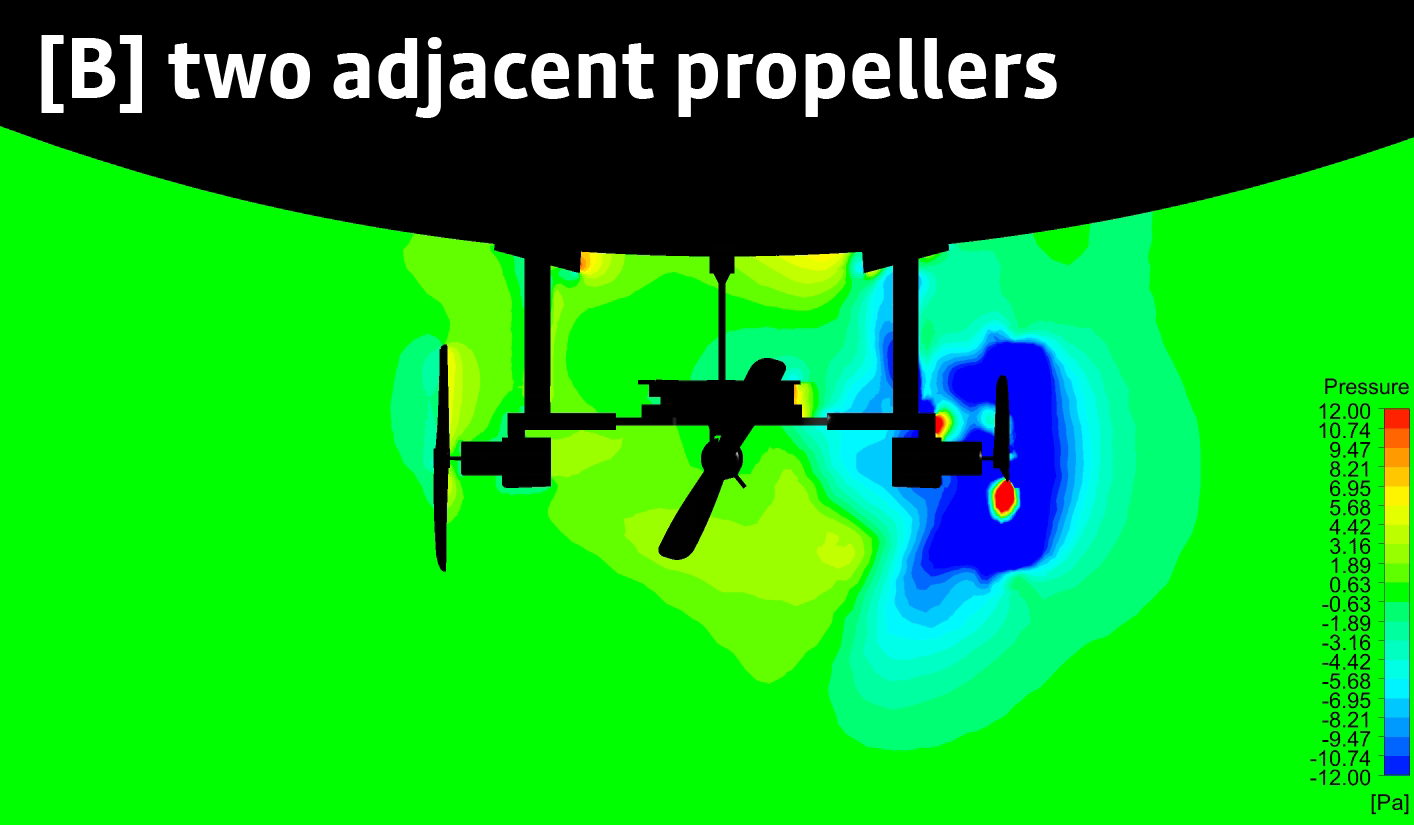}
    \end{subfigure}
    \begin{subfigure}{0.49\linewidth}
        \centering
        \includegraphics[width=\linewidth]{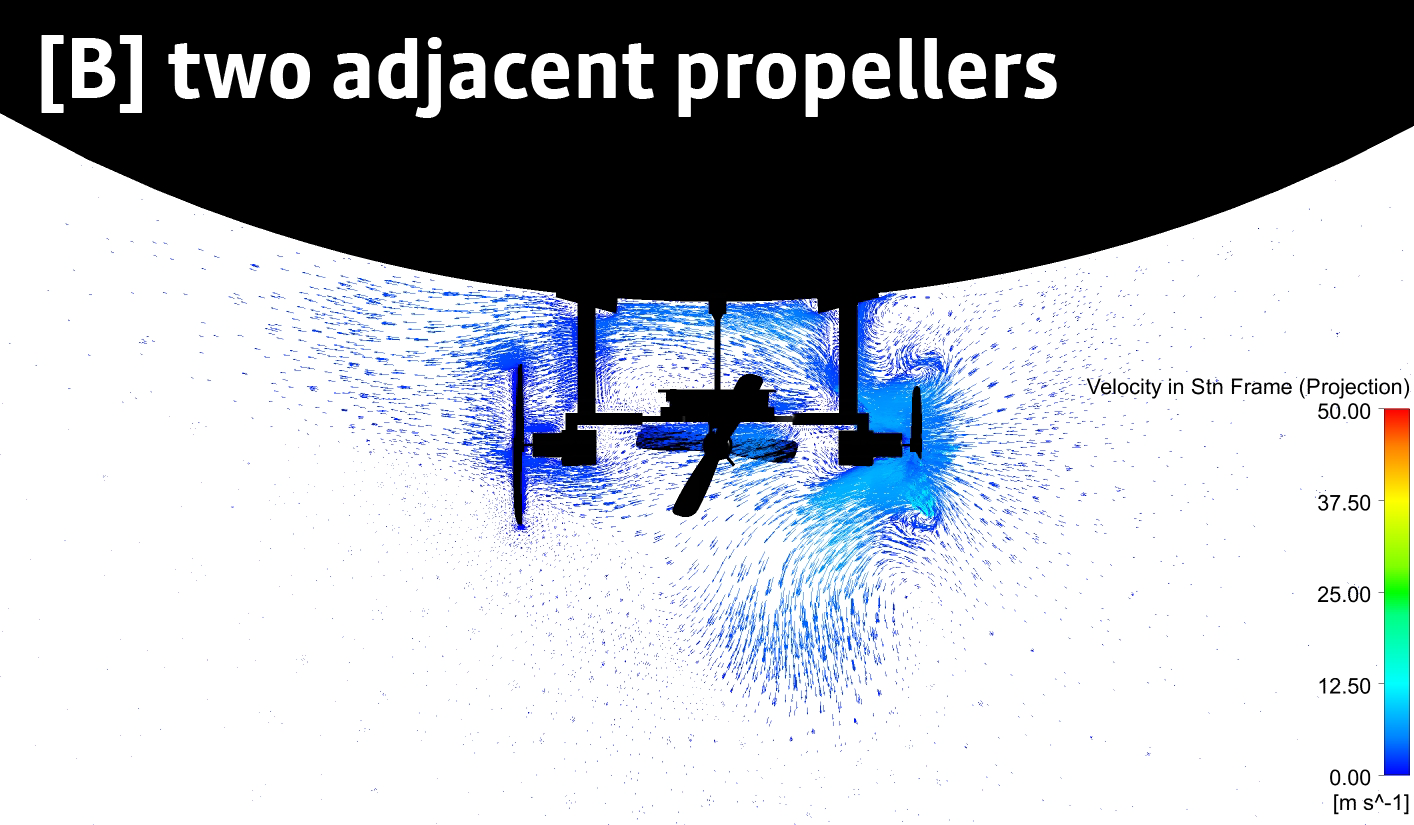}
    \end{subfigure}
    \begin{subfigure}{0.49\linewidth}
        \centering
        \includegraphics[width=\linewidth]{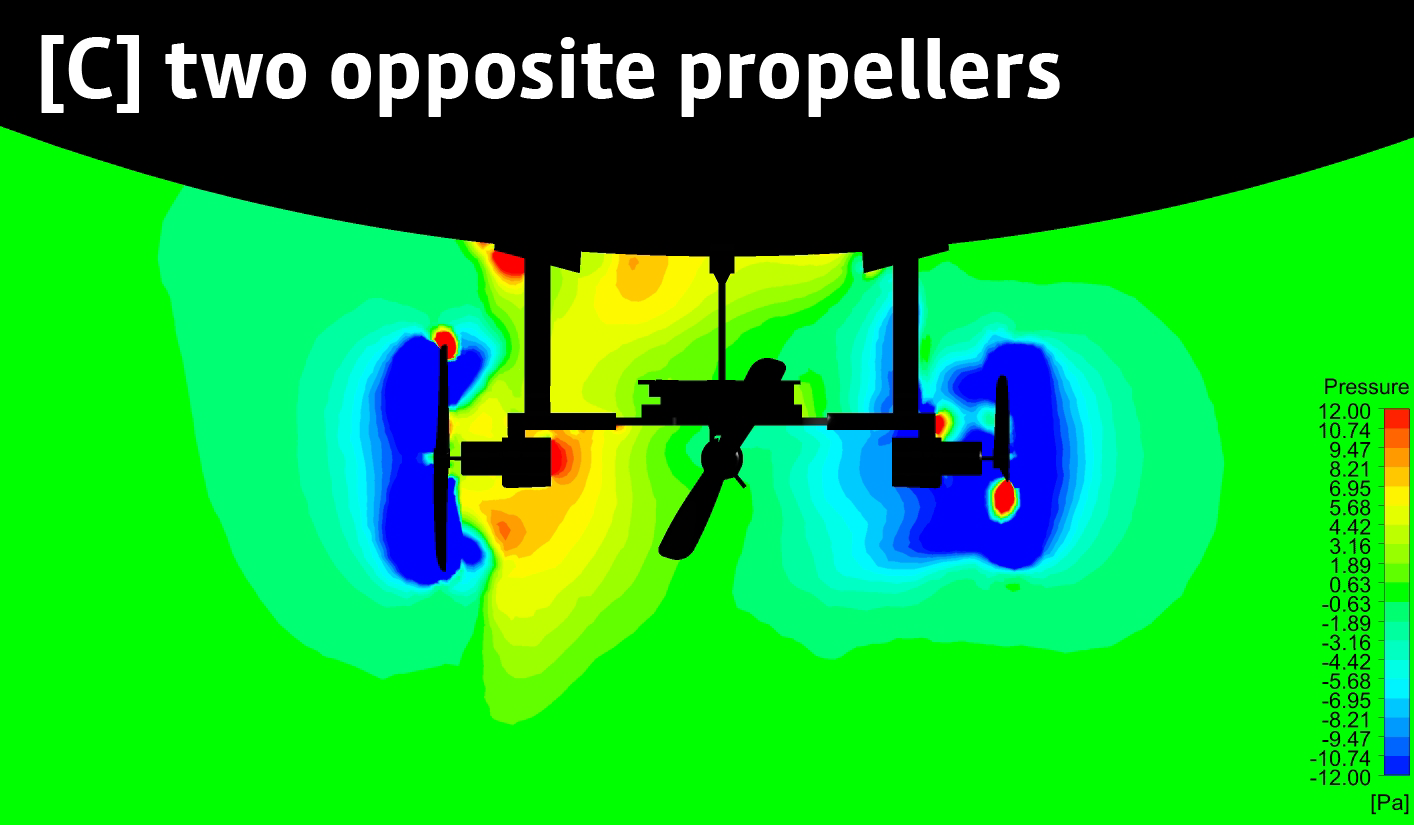}
    \end{subfigure}
    \begin{subfigure}{0.49\linewidth}
        \centering
        \includegraphics[width=\linewidth]{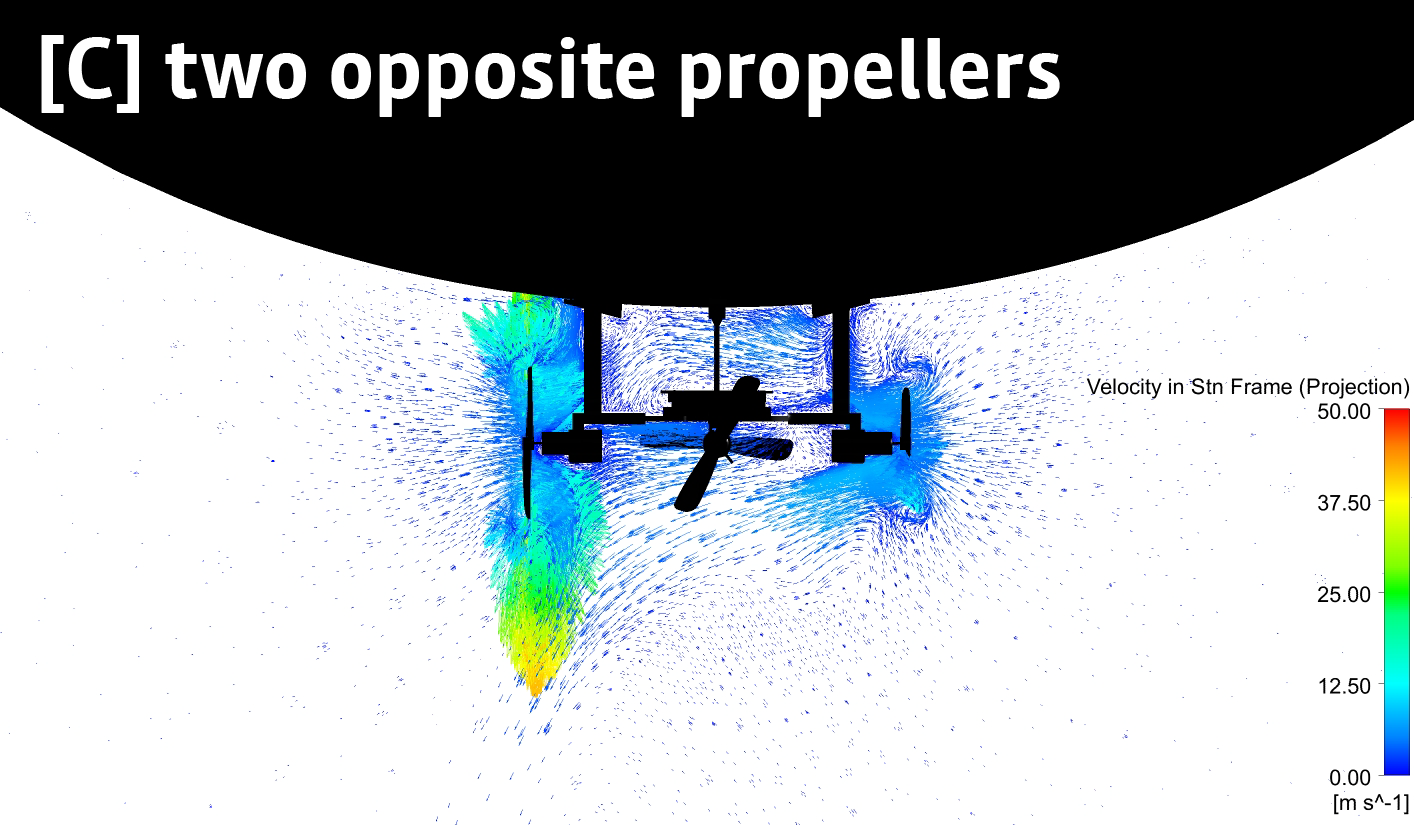}
    \end{subfigure}
    \begin{subfigure}{0.49\linewidth}
        \centering
        \includegraphics[width=\linewidth]{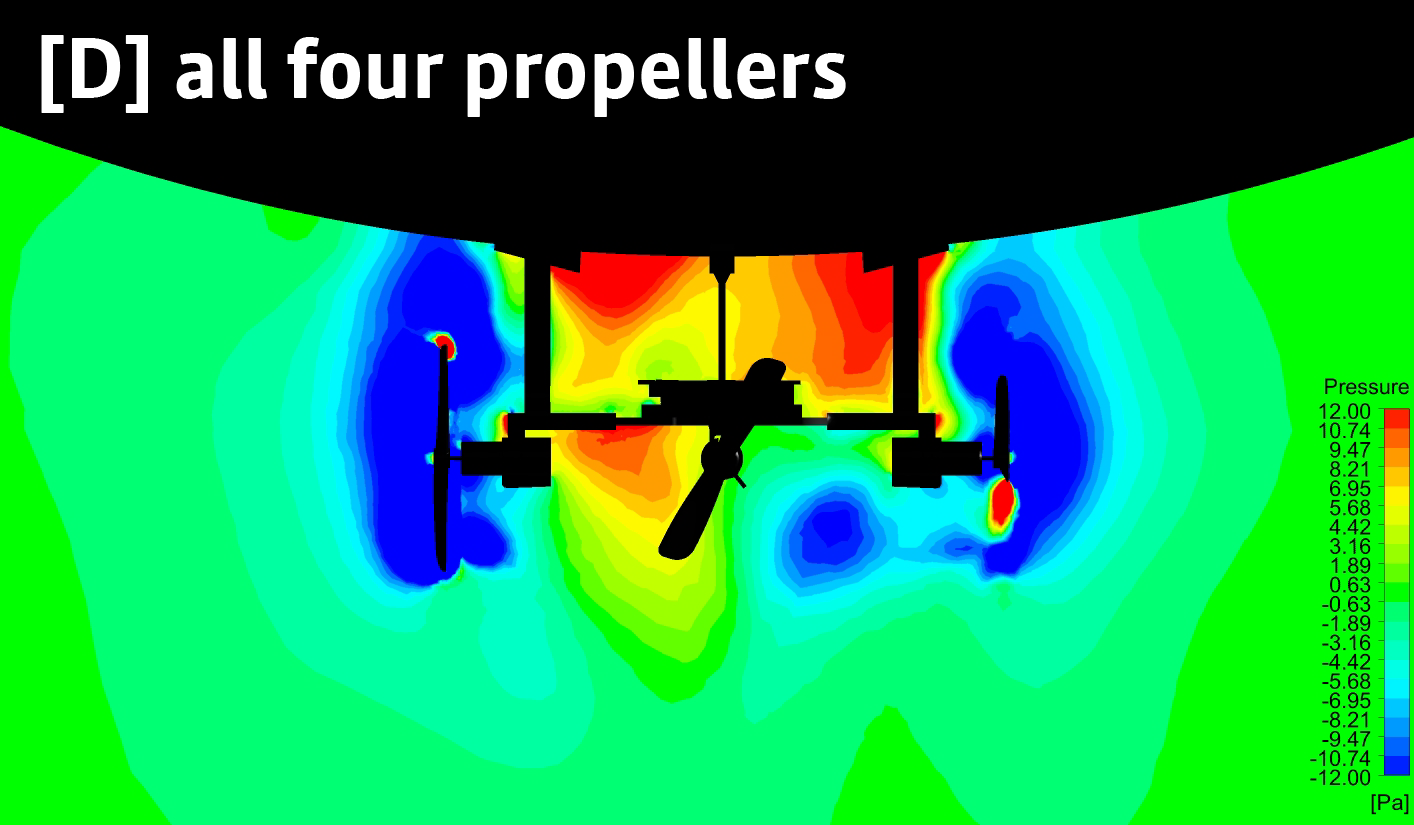}
    \end{subfigure}
    \begin{subfigure}{0.49\linewidth}
        \centering
        \includegraphics[width=\linewidth]{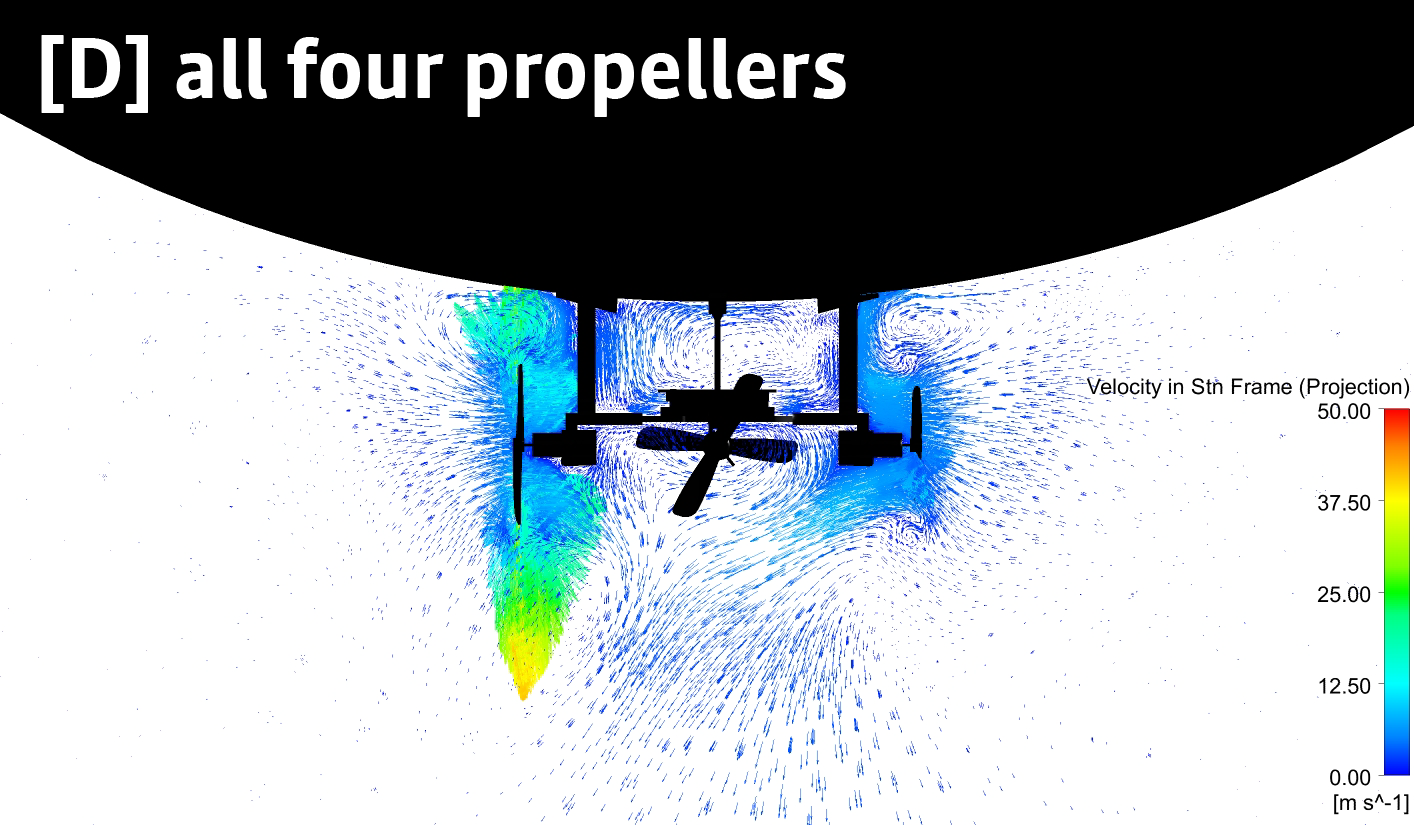}
    \end{subfigure}
    \caption{Pressure contours (left) and flow visualization (right) around \beavis with the varying number of motors running at full thrust.}
    \label{fig:CFD-flow-pressure}
\vspace{-15pt}
\end{figure}
The flow analysis is carried out using a hybrid mesh in Ansys Fluent~\cite{ANSYS}. The CFD simulations include the complete geometry of the platform (including the balloon) as a model, along with fluid domains for air and helium (inside the balloon) to take into account the changes in buoyancy due to the changes in the local flow conditions. A structured mesh is created to resolve the boundary layer over the propellers, while the rest of the domain is populated with unstructured mesh to discretize the rest of the flow domain with approximately 5\,M cells. Propeller rotation is modeled using a rotating frame of reference for all 4 propellers. 
Airflow is allowed to evolve for a total duration of 10\,s, by which the solution reaches a statistically steady state. A time step of 0.005\,s (total steps 10/0.005 = 2000) is used for the transient simulation, with 10 iterations per time step to achieve the required level of convergence. The solution is monitored using flow residuals and forces on different parts of the system. The solution is assumed to be converged when the lift on the quadrotor and balloon system reaches a statistically steady state and the flow residuals are below $5e^{-5}$ for all time steps. The resulting solution files are then post-processed in Ansys CFD-Post to visualize the flow contours and velocity vectors. The results from this analysis are presented in Fig.~\ref{fig:CFD-flow-pressure}. Four cases are considered: single, two adjacent, two opposite, and all four propellers running. 

\fakepar{Single propeller} It is seen that there are clear low-pressure regions formed both upstream and downstream around the rotating blade. The low-pressure region is formed upstream of the blade due to the formation of the wake (the region behind the body around which the fluid flows), while the one downstream is formed because of the air it displaces. There are also small, high-pressure pockets present near the balloon. These occur when the surrounding air is pushed into the mechanical structure. As a result of the low-pressure regions, there is a tendency for the platform to be pushed downwards. However, the presence of regions of comparatively high pressure around the wake causes this effect to be quite small. Dynamic buoyancy effects also influence the total upward force acting on the system and therefore, as a whole, this has a negligible effect on its motion.

\fakepar{Two adjacent propellers} There is mixing between the wakes which causes an increase in the local flow velocity near the rotating propeller blades. This effect of wake interaction causes a downward movement of the system. But in this case, the effect of dynamic buoyancy does not counteract the effect of the wake. There is a significant impact on the higher-pressure regions created by restricted air that try to push the system upwards. Therefore, it is observed that the entire system only moves downwards weakly.

\fakepar{Two opposite propellers} As expected, the interaction of the two wakes is non-existent, with the wakes from each propeller being restricted to around the periphery of the rotating blades. However, two very interesting flow phenomena occur here: (a) the two opposite wakes create a circulation that, on one side, generates extremely high-pressure regions (near the body of the platform and balloon); (b) on the other side only low-pressure pockets create (near the motor and propeller). These pockets have high velocities at the tips of the rotating propellers, and above and below them as well. This implies that when there are no wake interactions between adjacent propellers, the impact of buoyancy seems to be lowered. Thus the higher-pressure regions and the dynamic buoyancy effect have a lesser impact compared to the low-pressure wake regions. Therefore, the entire platform is pushed downwards.

\fakepar{All four propellers} We can observe very high pressure due to the interaction of all the wakes with the platform and balloon individually as well as due to their combined effect. We can also see recirculating regions in between the platform and the balloon, as well as the presence of the additional regions above and below the propellers that have the highest velocity. This is similar to what is observed when two opposite propellers are turned on. The impact of the mixing of adjacent wakes reduces the overall effect of the low-pressure wake region. This is caused by the presence of two such wake regions (instead of one) and the presence of a mass of air that is essentially ``trapped'' in between the rotors and the platform. Therefore, the combined effect of this trapped mass of air along with the dynamic buoyancy causes the overall acceleration of the platform to be upwards. This is a very interesting phenomenon because while all the other cases make the overall system move downwards, the four-propeller case makes it moves upwards. This intermixing of flows is precisely what we utilize to move \beavis vertically.
\vspace{-15pt}
\section{Flight Controller}
\label{sec:controller}
The objective of the flight control system is two-fold: (a) mainly to maintain a desired altitude and yaw (heading) angle and (b) to steer the drone in the lateral plane.
\vspace{-5pt}
\subsection{Nonlinear flight dynamics model}
\begin{figure}[!tb]
    \centering
        \includegraphics[width=0.9\linewidth]{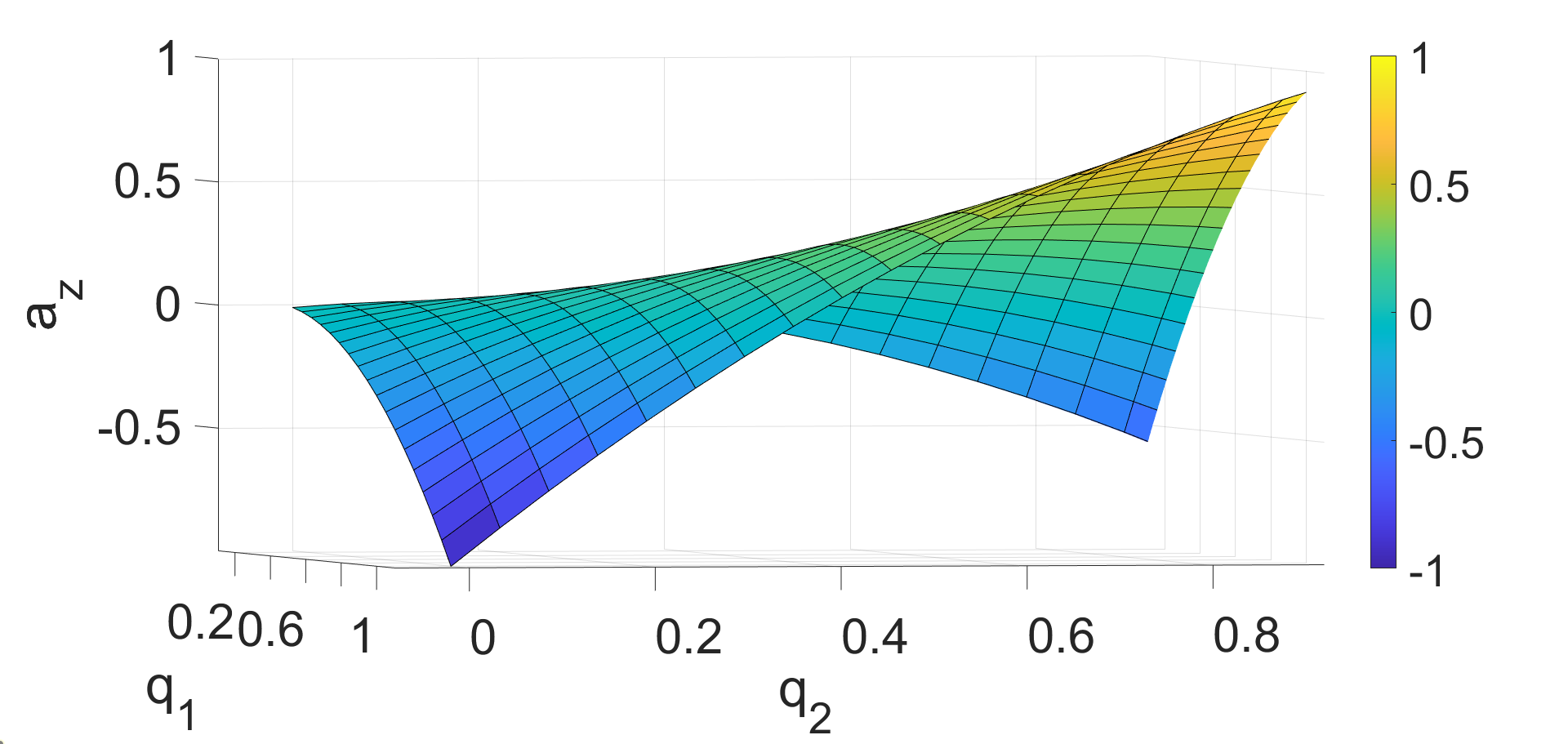}
        \vspace{-5pt}
    \caption{3D plot showing the relationship $q_1q_2 - (q_1 - q_2)^{2}$ between $q_1$, $q_2$ and $a_z$ of the system.}
    \label{fig:3D-plot}
\vspace{-5pt}
\end{figure}

The flight control dynamics of \beavis can be derived in the earth-fixed inertial frame as shown in \ref{fig:motor-mount} via the Euler-Lagrange equations of motion as,
\begin{equation}
\vspace{-2pt}
  \begin{rcases}
   \begin{aligned}
\dot{x}&=v_x, ~ \dot{y}=v_y, ~ \dot{z}=v_z, \\
\dot{v}_x&=a_x=(M_3+M_4-M_1-M_2)/m-C_dv_x, \\
\dot{v}_y&=a_y=(M_3+M_2-M_1-M_4)/m-C_dv_y, \\
\dot{v}_z&=a_z=C_z(q_1q_2-(q_1-q_2)^2)-C_dv_zr, \\
\dot{\psi}&=\omega_z,~~ \dot{\omega}_z=\tau_z=C_{\psi}(q_1-q_2),
\label{dynamics}
   \end{aligned}
 \end{rcases}
\vspace{-5pt}
\end{equation}
where $x,y,z,v_x,v_y,v_z$ denote the position and velocity and $\psi,\omega_z$ denote the yaw angle and yaw angular rate. $M_i$ denotes the $i^{\text{th}}$ motor thrust, $q_1:=M_1+M_3,q_2:=M_2+M_4$ and $C_d,C_{\psi},C_{z}$ denote the aerodynamic coefficients, and $m$ denotes the mass. $\dot{x}$ is the time derivative of $x$. Note that the pitch and roll dynamics are ignored due to their natural stability. The coupling between lateral and angular accelerations and control thrusts is derived based on the CFD analysis. As it was shown that when all four motors are active, there is a positive lift and when only one pair of opposite motors are active there is a negative lift. The $\dot{v_z}$ equation is derived based on this relation. The nonlinear relation has been derived based on CFD (Fig.~\ref{fig:CFD-flow-pressure}) and flight data (Fig.~\ref{fig:3D-plot}). 
We can see that the equations are nonlinear and therefore traditional control design may not be effective in stabilizing or steering \beavis. 

\subsection{Nonlinear control design} 
\begin{figure}[!tb]
    \centering
    \includegraphics[width=0.9\linewidth]{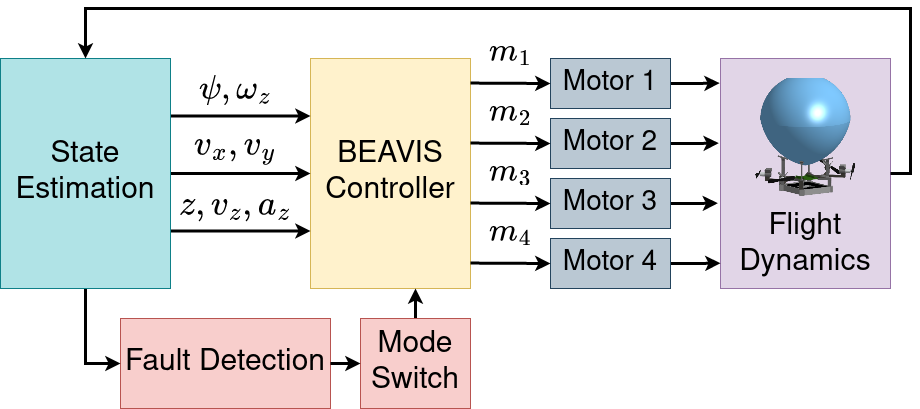}
    \caption{Block diagram of the model-free controller.}
    \vspace{-10pt}
    \label{fig:model-free-controller}
\end{figure}

The relation between the angular and lateral accelerations and control inputs can be written as, \begin{equation}
\vspace{-5pt}
(a_x,a_y,a_z,\tau_z) :=F(M_1,M_2,M_3,M_4),
\end{equation}
where $F$ is derived from the right-hand side of Eq.~\eqref{dynamics}. One approach to controlling \beavis is to invert this relation based on the requirement of accelerations. However, this may be computationally ineffective, moreover, the control bandwidth required for altitude and yaw is higher than that of lateral steering, and the sensing modalities and consequently sensing rate and error are different as well. Therefore, a hierarchical approach is taken (see Fig.~\ref{fig:model-free-controller}) as below. 

\noindent\textbf{STEP 1:} First, a controller is designed for only altitude and yaw. Denote $(a_z,\tau_z)=F_1(q_1,q_2)$, where $F_1$ is obtained from Eq.~\eqref{dynamics}. Let $z_d$ and $\psi_d$ denote the desired values. Then, $q_1,q_2$ are assigned as follows,
    \begin{equation}
    \vspace{-4pt}
  \begin{rcases}
   \begin{aligned}
    a_z=K_p(z-z_d)-K_d(v_z-\dot{z}_d)-K_i\int(z-z_d)dt, \\
    \tau_z=K_{p}(\psi-\psi_d)-K_d(\omega_z-\dot{\psi}_d), \\
    (q_1,q_2)=F_1^{-1}(a_z,\tau_z),
    \label{z control}
       \end{aligned}
 \end{rcases}
 \vspace{-2pt}
\end{equation}
where $F_1^{-1}$ is obtained via standard methods (Newton-Raphson iterative inversion). The terms $K_{*}$ denote control gains. The control law so far assumes that $M_1=M_3,M_2=M_4$.
 
\noindent\textbf{STEP 2:} Given desired lateral position $x_{d},y_{d}$, $(M_3-M_1)$ and $(M_4-M_2)$ are varied while maintaining $q_1,q_2$ as derived in STEP 1 as follows, 
      \begin{equation}
    \vspace{-8pt}
  \begin{rcases}
   \begin{aligned}
    a_x&=K_p(x-x_d)-K_d(v_x-\dot{x}_d)-K_i\int(x-x_d)dt, \\
    a_y&=K_p(y-y_d)-K_d(v_y-\dot{y}_d)-K_i\int(y-y_d)dt, \\
        M_1&=(q_1-a_x+a_y)/2,~M_3=(q_1+a_x+a_y)/2, \\
    M_2&=(q_2-a_x-a_y)/2,~ ~ M_4=(q_2+a_x-a_y)/2,
        \label{lateral control}
     \end{aligned}
 \end{rcases} 
\end{equation}
\subsection{Fault tolerant control}

\begin{algorithm}[!tb]
\footnotesize
	\DontPrintSemicolon
    \SetAlgoLined
    \KwResult{Determine $FAILED_{mi}, \forall i\in [1,4]$}
    Set $\psi_{desired}$, $K_{\omega_{z}}$, $K_{a_{z}}$
    
    Measure $\psi_{true}$, $\omega_{true}$, $a_{z_{true}}$, $v_{x_{true}}$, $v_{y_{true}}$

    Estimate $\omega_{z_{exp}}$, $a_{z_{exp}}$, $v_{x_{exp}}$, $v_{y_{exp}}$ from Equation~\eqref{dynamics}

    Calculate $ e_{\psi} = \psi_{true} - \psi_{desired}$

    Calculate $ e_{\omega_{z}} = \omega_{z_{true}} - \omega_{z_{exp}}$, $ e_{a_{z}} = a_{z_{true}} - a_{z_{exp}}$, $ e_{v_{x}} = v_{x_{true}} - v_{x_{exp}}$, $ e_{v_{y}} = v_{y_{true}} - v_{y_{exp}}$

    \If{ $\mid{e_{\psi}}\mid > 0$ }{ 
        \If{$\mid{e_{\omega_{z}}}\mid > K_{\omega_{z}}$ \textbf{and} $\mid{e_{a_{z}}}\mid > K_{a_{z}}$ }
        {
            \If{ $\omega_{z} < 0$}
            {
                \uIf{$v_{y_{true}} \neq v_{y_{exp}}$ \textbf{and} $v_{y_{true}} > 0$}
                {
                    $FAILED_{m1} = True$
                }
                \uElseIf{$v_{y_{true}} \neq v_{y_{exp}}$ \textbf{and} $v_{y_{true}} < 0$}
                {
                    $FAILED_{m2} = True$                
                }
                \Else{
                    $FAILED_{mi} = False, \forall i\in [1,2]$              
                }
            }
            \Else{
                \uIf{$v_{y_{true}} \neq v_{y_{exp}}$ \textbf{and} $v_{y_{true}} > 0$}{
                    $FAILED_{m4} = True$
                }
                \uElseIf{$v_{y_{true}} \neq v_{y_{exp}}$ \textbf{and} $v_{y_{true}} < 0$}{
                    $FAILED_{m3} = True$                
                }
                \Else{
                    $FAILED_{mi} = False, \forall i\in [3,4]$ 
                }
            }
            }
        \Else{
            $FAILED_{mi} = False, \forall i\in [1,4]$            
        }
        }
        \Else{
        End
        }
\caption{Rotor fault detection algorithm}
\label{algo:fault-detection}
\end{algorithm}%
To make the controller robust, we design an algorithm to detect rotor failure and isolate the failed rotor. Having done so, the control system is reconfigured to steer the drone to the desired location after descending. \textcolor{black}{A fault is detected whenever the difference in expected and actual behaviour exceeds a threshold, set to prevent misdiagnosis caused by external disturbances. This is a function of the operating environment. We set this to 1-1.5s in our trials.}

\noindent\textbf{STEP 1: Fault detection and isolation} As a first step, the signals from the IMU are analyzed along with the commanded thrust inputs, and a deviation from the expected behaviour (based on the model Eq.~\eqref{dynamics}) is used as an indicator of fault. To distinguish from external disturbances such as wind gusts, the signals are analyzed over a sufficient duration and compared against set thresholds $K_{\omega_{z}},K_{a_{z}}$. Over several flight observations, we notice that a rotor failure results in anomalous spin due to an imbalance in the yaw moment. Further, there is also a resulting loss in altitude. The direction of spin indicates which pair of motors (i.e. in $q_1$ or $q_2$) are faulty. Next, the direction of lateral displacement indicates the particular motor in the pair which is to be isolated as shown in Algorithm \ref{algo:fault-detection}. 

\noindent\textbf{STEP 2: Control switching}
In case a failed rotor has been identified, then the control objective is modified to descend and steer the drone toward a point of recovery. To achieve this, the rotor adjacent to the failed one which is farther is also switched off, and the drone is controlled in the lateral plane using only two adjacent rotors (like a typical blimp configuration). However, the unique feature of \beavis as shown in Fig.\ref{fig:CFD-flow-pressure} is that it descends when only two adjacent rotors are active. Therefore, the \texttt{Z}-axis control objective is automatically achieved while steering in the lateral plane. However, since the control is under-actuated, the desired lateral acceleration is determined and the desired yaw angle is derived from it, and not independently controlled as in normal operation, which is,
 \begin{equation}
  \begin{rcases}
   \begin{aligned}
\psi_d&=\arctan(a_y,a_x), \\
M_3&=(||a_x+a_y||+\tau_z)/2, \\
M_2&=(||a_x+a_y||-\tau_z)/2,
     \end{aligned}
 \end{rcases} 
 \vspace{-5pt}
\end{equation}
where $(a_x,a_y,\tau_z)$ are derived from \eqref{lateral control}. Assume without loss of generality that $M_1$ has failed, in which case $M_4$ is also turned off. 
\begin{figure*}[!tbh]
    \centering
    \begin{subfigure}{0.245\linewidth}
        \centering
        \includegraphics[width=\linewidth]{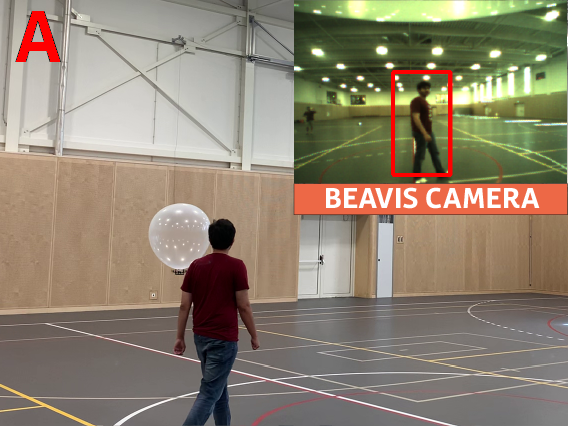}
    \end{subfigure}
    \begin{subfigure}{0.245\linewidth}
        \centering
        \includegraphics[width=\linewidth]{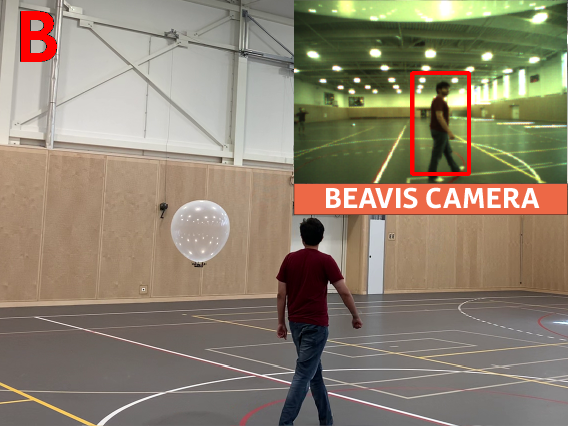}
    \end{subfigure}
    \begin{subfigure}{0.245\linewidth}
        \centering
        \includegraphics[width=\linewidth]{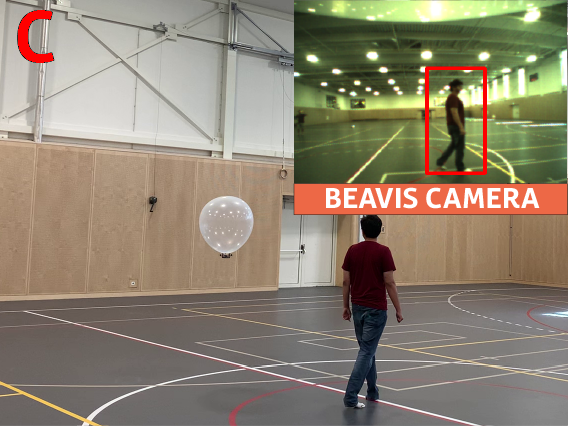}
    \end{subfigure}
    \begin{subfigure}{0.245\linewidth}
        \centering
        \includegraphics[width=\linewidth]{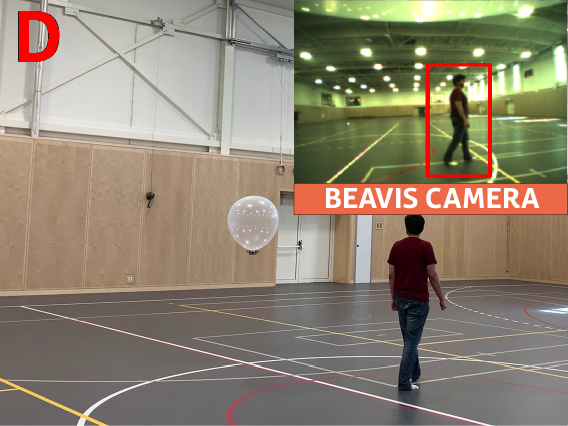}
    \end{subfigure}
    \caption{Using \beavis for crowd monitoring and people tracking with a front-facing RGB camera~\cite{video-link}.}
    \label{fig:camera-tracking}
\vspace{-5pt}
\end{figure*}
\vspace{-5pt}
\section{Evaluation}
\label{sec:eval}
We now provide a detailed experimental evaluation of the \beavis prototype (Video~\cite{video-link}). We use a reference payload ranging from 55\,g - 60\,g (almost half of the total weight of \beavis) in our experiments to simulate a sensor or a radio payload. 
\textcolor{black}{Proximity to an airport and regulations (due to inherent risks) for experimental UAVs (including helium balloons) deterred us from experimenting outside. All the experiments are carried out indoors and the prototype is fabricated keeping this limitation in mind. We tried to maintain diversity in our testing environment by spanning it across multiple locations as seen in our video~\cite{video-link}.} 

In total, we conducted 10 separate flight tests using distinct similarly sized balloons each time. The balloons are filled with commercially available helium gas and neutral buoyancy is achieved by adding weight to the ballast. We log all of the flight data to a ground station using the wireless radio link present on \beavis. This resulted in over 30 flight logs with sensed parameters like acceleration, velocities, and altitude being recorded along with the battery voltage. This data was also used in the development of our nonlinear data-driven autonomous controller. We verify all the degrees of freedom that we could achieve as well as characterize the system performance metrics as discussed in \ref{sec:system-metrics}. We implement two applications that showcase the potential utility of our platform in real-world sensing use cases.
\begin{figure}[!tb]
    \centering
    \begin{subfigure}{0.49\linewidth}
        \centering
        \includegraphics[height=\linewidth, width=\linewidth]{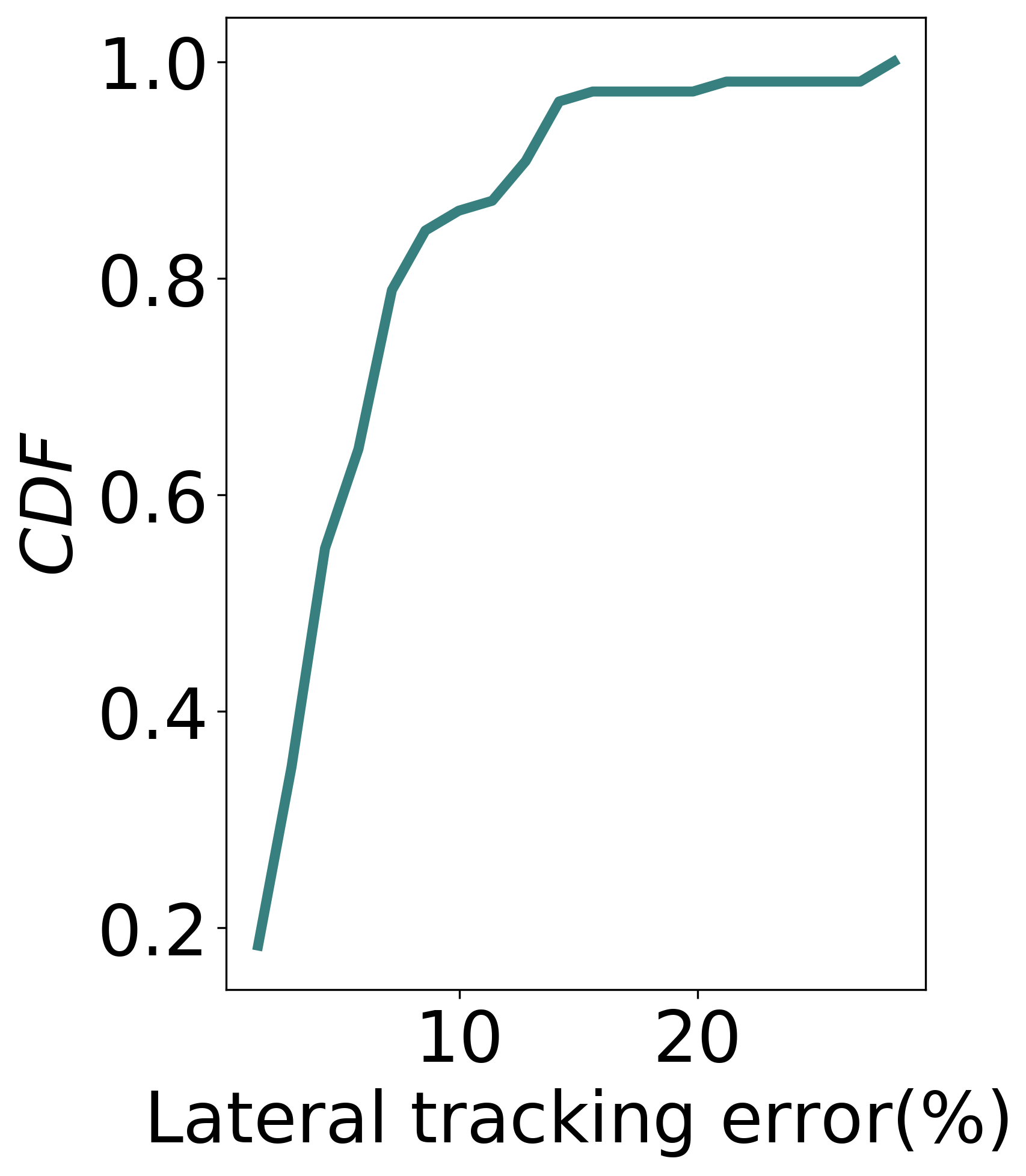}
    \end{subfigure}
    \begin{subfigure}{0.49\linewidth}
        \centering
        \includegraphics[height=\linewidth, width=\linewidth]{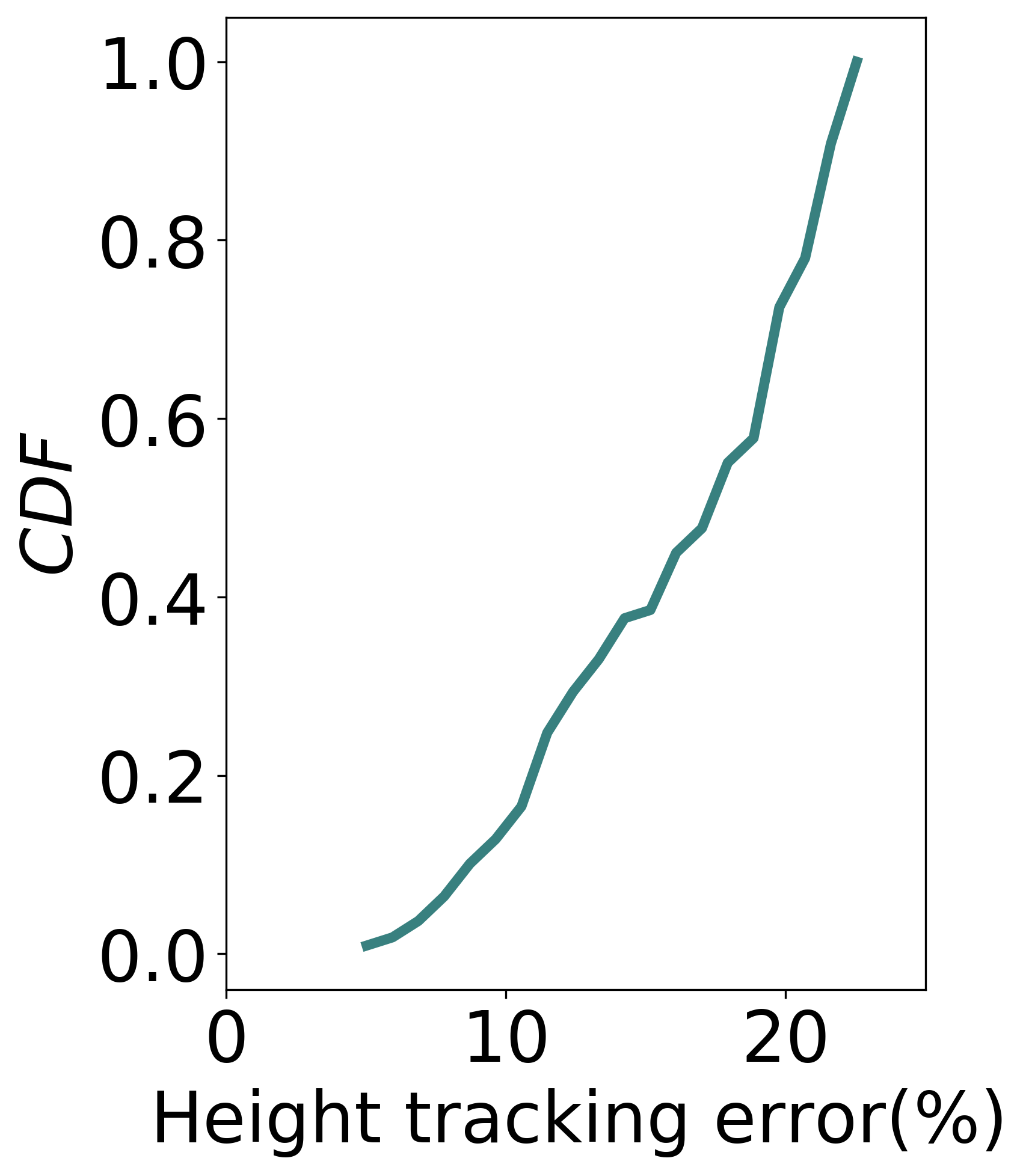}
    \end{subfigure}
    \caption{CDF of tracking error percentage (lateral and height) for target tracking with a front facing camera.}
    \label{fig:CDF-tracking}
\vspace{-10pt}
\end{figure}
\vspace{-5pt}
\subsection{Application Scenarios}
\label{sec:app-eval}
\fakepar{People and object tracking} People and object tracking application is a combination of high endurance and high mobility of the \beavis platform. Using Himax HM01B0-ANA front-facing color camera capable of capturing video at 60 frames per second (QVGA) with a 324x324 pixel resolution, we tracked a single person to showcase the agile flight control the platform has. We implemented a machine learning-based people tracking algorithm~\cite{OG-tracking} where \beavis tries to maintain the largest detected person in the center of the frame. Based on this input the platform calculates the new heading  (see Fig.~\ref{fig:camera-tracking}) as well as the lateral (\texttt{X}) and height (\texttt{Y}) error. The four frames show the quadrotor detecting a person in the frame and adjusting itself. With a bigger quadrotor, one can monitor the crowd on the ground.
Fig.~\ref{fig:CDF-tracking} shows the cumulative distribution of the tracking error while following a moving ground target. \textcolor{black}{The tracking error is measured from the centre of the frame to the centre of the detected human for both X and Y.} The ground target moved at an average speed of 100\,cm/s. The platform tracked the target for 2000\,cm. The results show that \beavis could maintain its lateral (horizontal) tracking error under 12.5\% with 90\% probability and its height (vertical) tracking error under 21\% with 90\% probability. 
\begin{figure}[!tb]
    \centering
    \begin{subfigure}{0.325\linewidth}
        \centering
        \includegraphics[width=\linewidth]{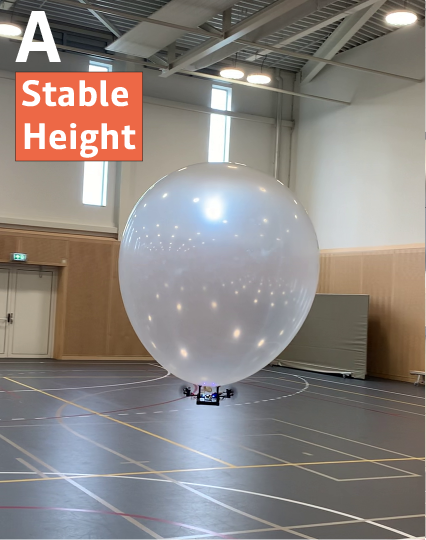}
    \end{subfigure}
    \begin{subfigure}{0.325\linewidth}
        \centering
        \includegraphics[width=\linewidth]{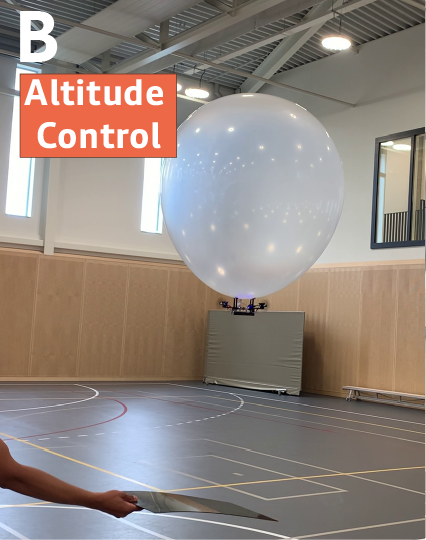}
    \end{subfigure}
    \begin{subfigure}{0.325\linewidth}
        \centering
        \includegraphics[width=\linewidth]{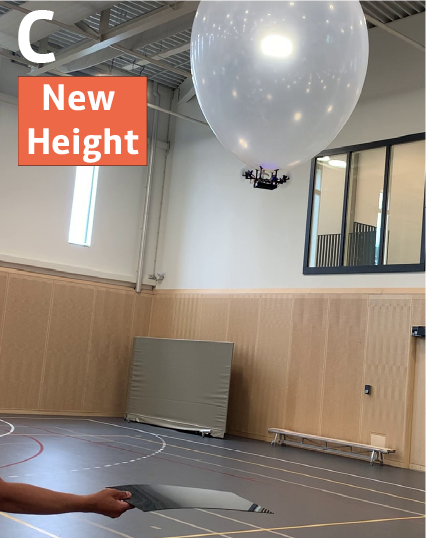}
    \end{subfigure}
    \begin{subfigure}{0.49\linewidth}
        \centering
        \includegraphics[width=\linewidth]{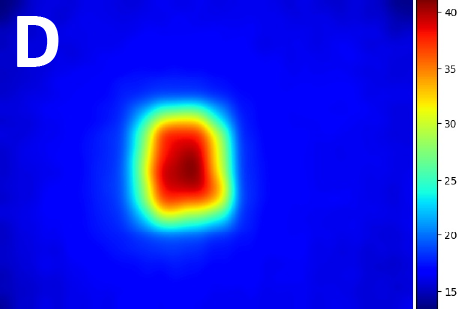}
    \end{subfigure}
        \begin{subfigure}{0.49\linewidth}
        \includegraphics[width=\linewidth]{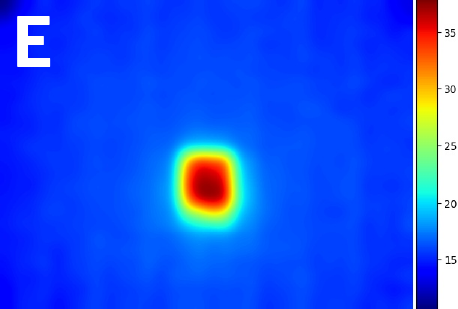}
        \centering
    \end{subfigure}
    \caption{\beavis response while flying at a stable height with a thermal camera (a), the sensed height is halved (b) and the platform corrects it (c). The thermal image without height adjustment (d) and after correcting it (e) using our altitude control~\cite{video-link}.}
    \label{fig:thermal-altitude-control}
\vspace{-15pt}
\end{figure}

\fakepar{City rooftop thermal imaging} Cities are gearing up to reduce energy in heating buildings.  Estimating the heat escape from the rooftops is an important indicator. This application was requested by a local municipality that wanted to quantify the wastage of heat by monitoring rooftops. To aid this, we use a very \textit{low pixel thermal camera} guaranteeing privacy. \beavis can help since it can hover over buildings for a long duration. However \beavis needs to automatically adjust its hovering height depending on the height of the buildings to avoid manual intervention. This is required to normalize the area of the rooftop in the thermal image and estimate heat escape. We demonstrate the application of capturing thermal images by using a downward-facing thermal camera module, MLX90640-D110. This is a 32 × 24 pixel, \ang{110} FOV, IR array thermal imaging camera which can measure temperature with a resolution of $\pm$\ang{1}\,C. We move \beavis around in a pre-determined trajectory and altitude is adjusted based on the height of the ground (read as building here) below. 

The platform adjusts itself using our altitude control algorithm to maintain a constant separation from the objects below. The time of flight sensor is used to estimate its altitude from the ground. To demonstrate, a black acrylic sheet is introduced in its FOV, the platform estimates that it is closer to the ground than desired and adjusts itself. The thermal image captured is also shown in Fig.~\ref{fig:thermal-altitude-control}.
\begin{figure}[!tb]
        \begin{subfigure}{\linewidth}
            \centering
            \includegraphics[width=0.7\linewidth]{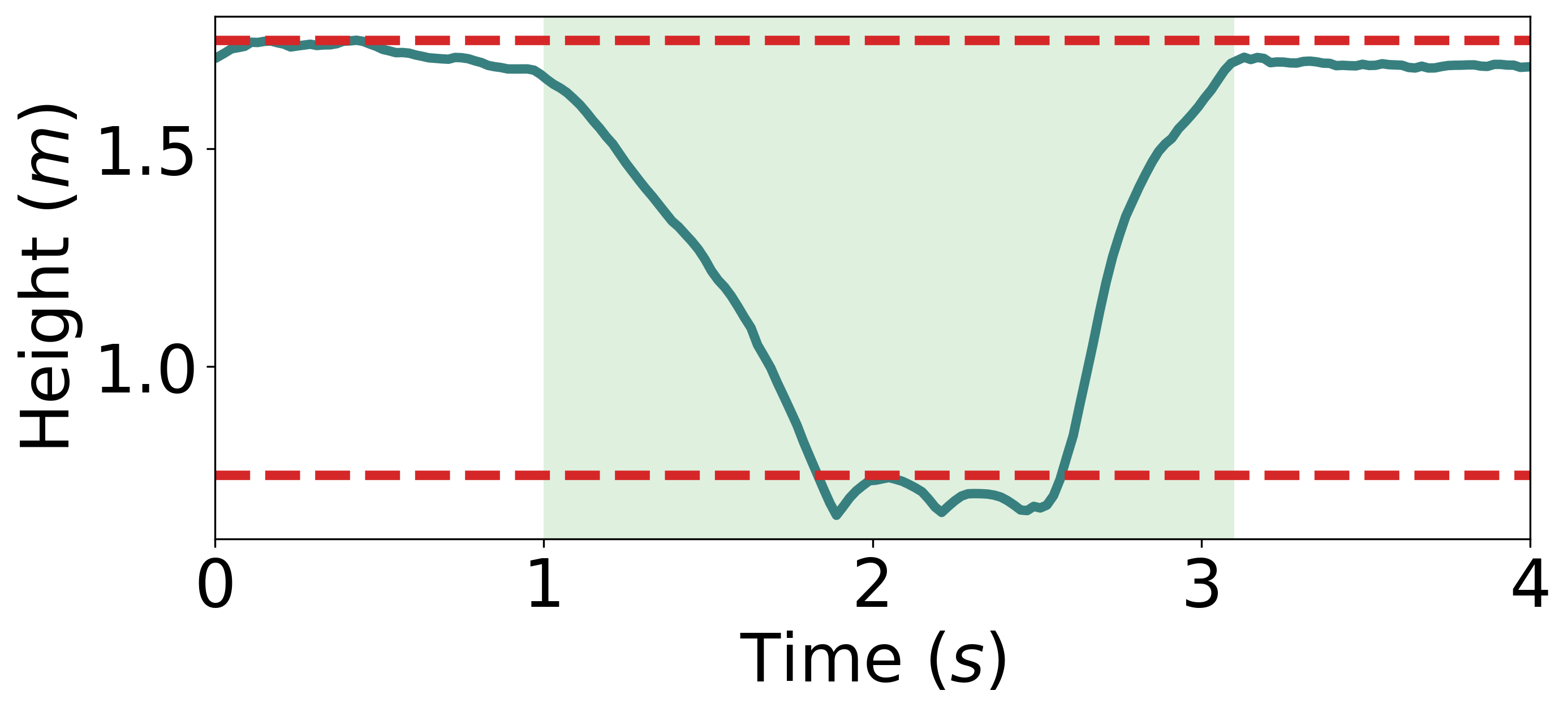}
        \end{subfigure}
        \begin{subfigure}{\linewidth}
            \centering
            \includegraphics[width=0.72\linewidth]{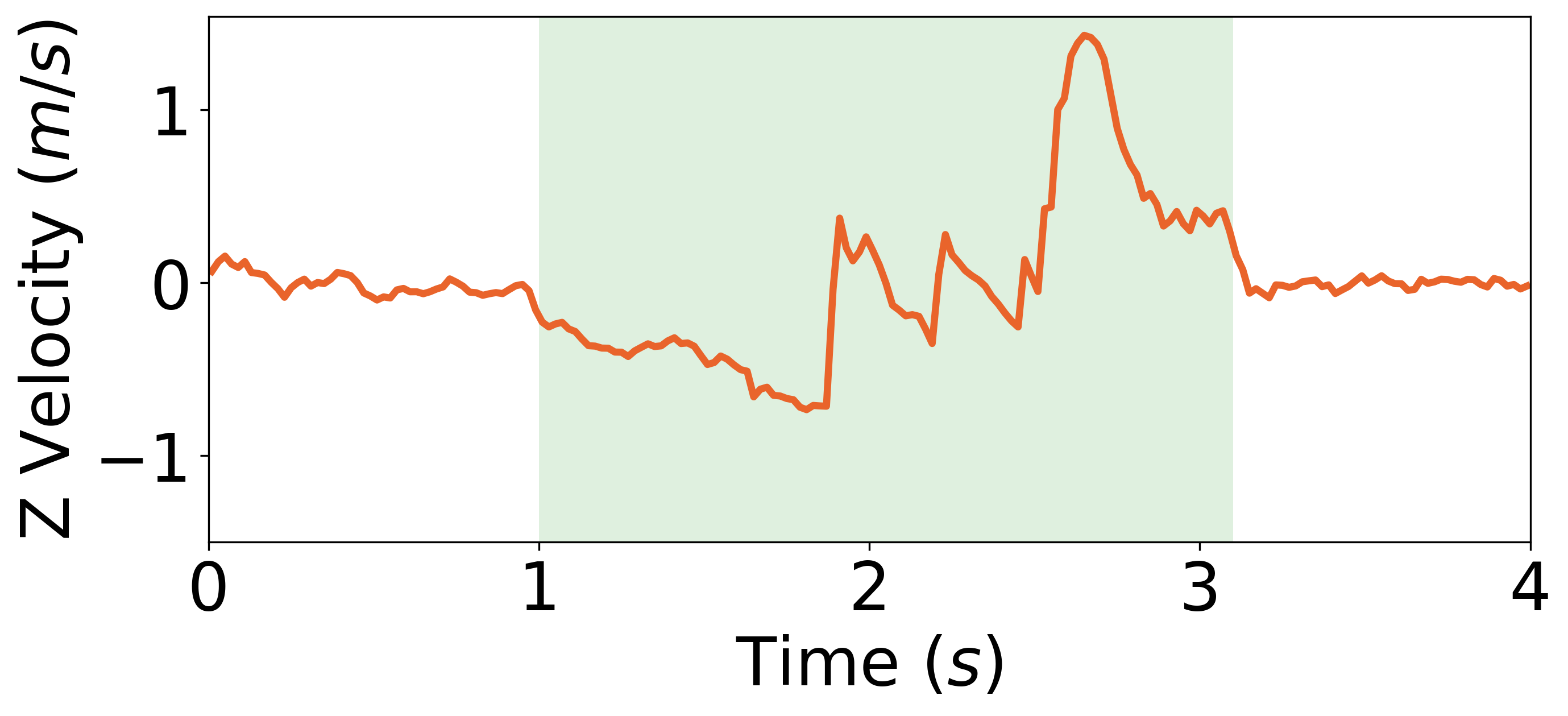}
        \end{subfigure}
    \caption{The variation of height (\texttt{z}) and vertical velocity $v_z$ when sensed height is halved using acrylic sheet.}
\label{fig:height-adjust-thermal}
\vspace{-5pt}
\end{figure}
\begin{figure}[!tb]
    \centering
        \includegraphics[width=0.85\linewidth]{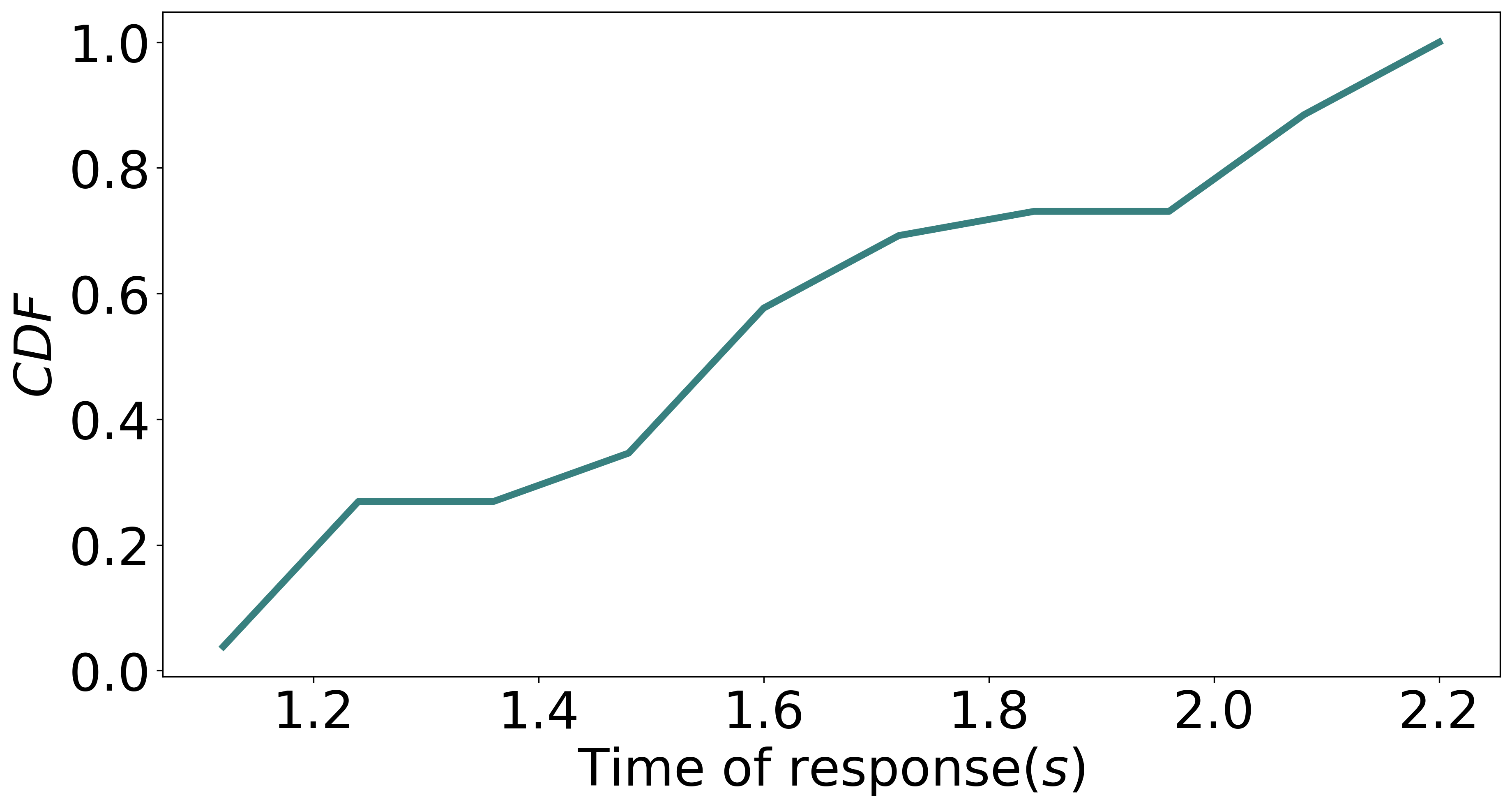}
    \caption{CDF of the speed of response of the platform with altitude control.}
    \label{fig:CDF-thermal}
\end{figure}

Further to evaluate the response of the platform, we also recorded the time it takes to adjust the height. This is crucial because if the \textcolor{black}{response time is low} then more houses can be covered in one flight. In our experiments, we introduce an acrylic sheet suddenly to reduce the height (separation) to less than half. In our trials, we set this height to 175\,cm which is reduced to 75\,cm. Fig.~\ref{fig:CDF-thermal} shows the response time of \beavis in terms of height (\texttt{z}) and vertical velocity $v_z$. The height of the platform is reduced from 1.75\,m before $t=0.5$\,s to 0.75\,m at $t=0.5$\,s. The platform then responds by readjusting itself to 1.75\,m separation. We record the time it takes to recover from this abrupt height change. 
Fig.~\ref{fig:CDF-thermal} shows the cumulative distribution of the response time for the platform to readjust its height after it detects itself to be at a lower height than desired. The graph shows that \beavis readjusts itself within 2.1\,s in 90\% of the cases showcasing the applicability of \beavis.
\vspace{-8pt}
\subsection{System Performance}
\label{sec:system-metrics}
\begin{figure}[!tb]
    \centering
    \begin{subfigure}{0.29\linewidth}
        \centering
        \includegraphics[width=\linewidth]{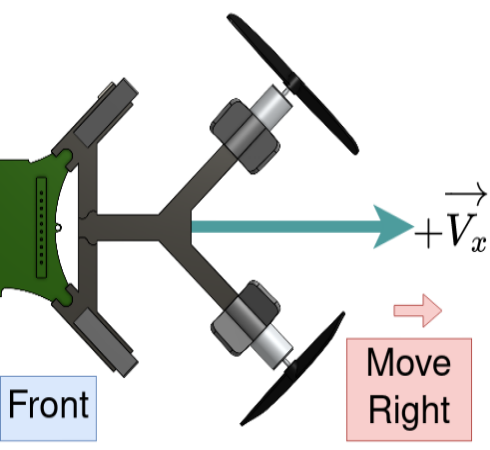}
    \end{subfigure}
    \begin{subfigure}{0.70\linewidth}
        \centering
        \includegraphics[width=\linewidth]{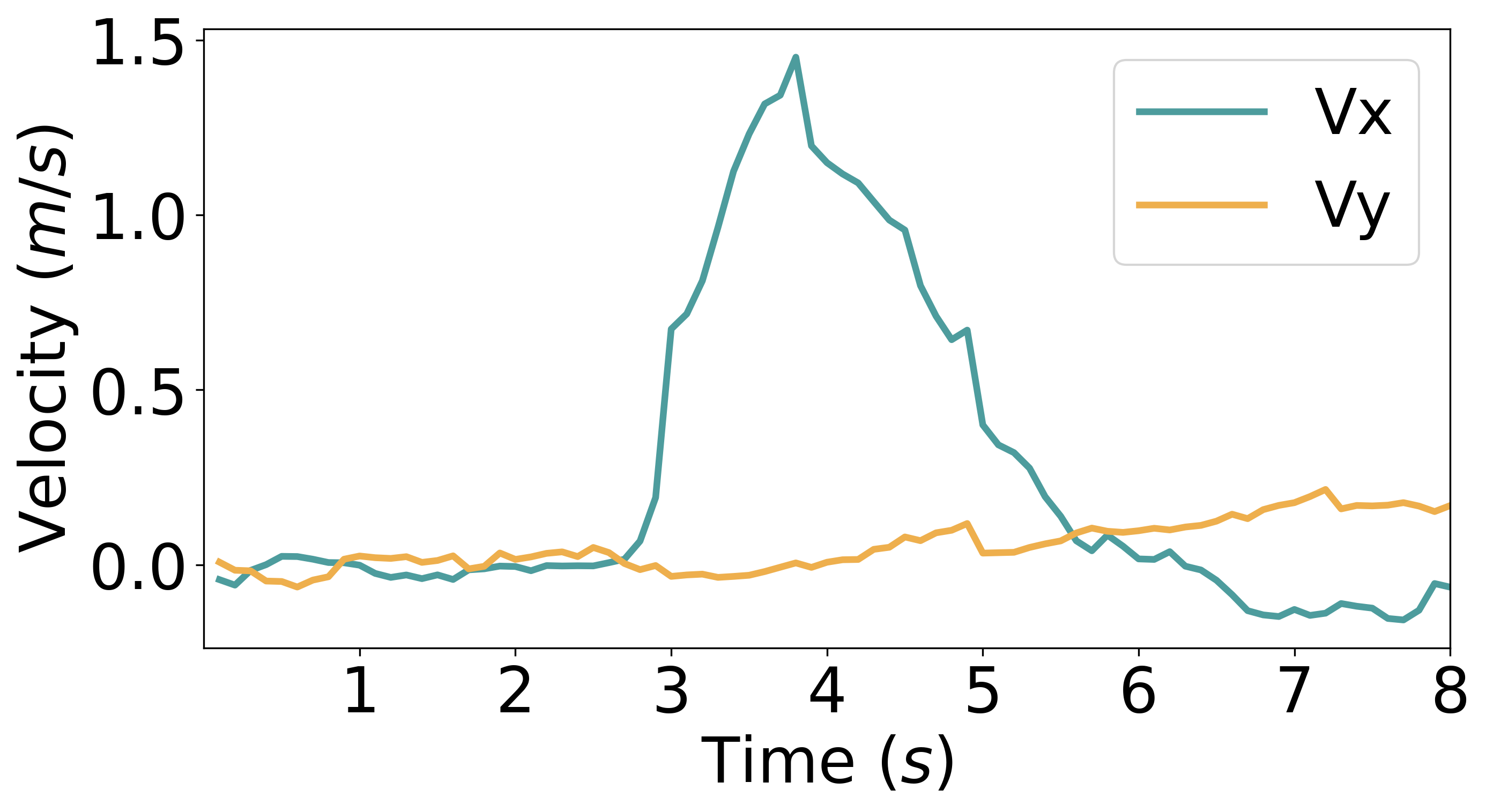}
    \end{subfigure}
    \begin{subfigure}{0.29\linewidth}
        \centering
        \includegraphics[width=\linewidth]{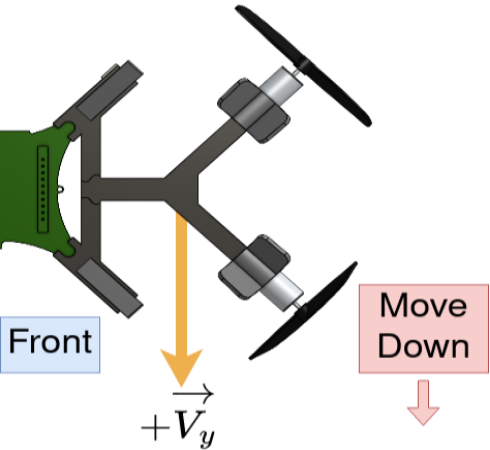}
    \end{subfigure}
    \begin{subfigure}{0.7\linewidth}
        \centering
        \includegraphics[width=\linewidth]{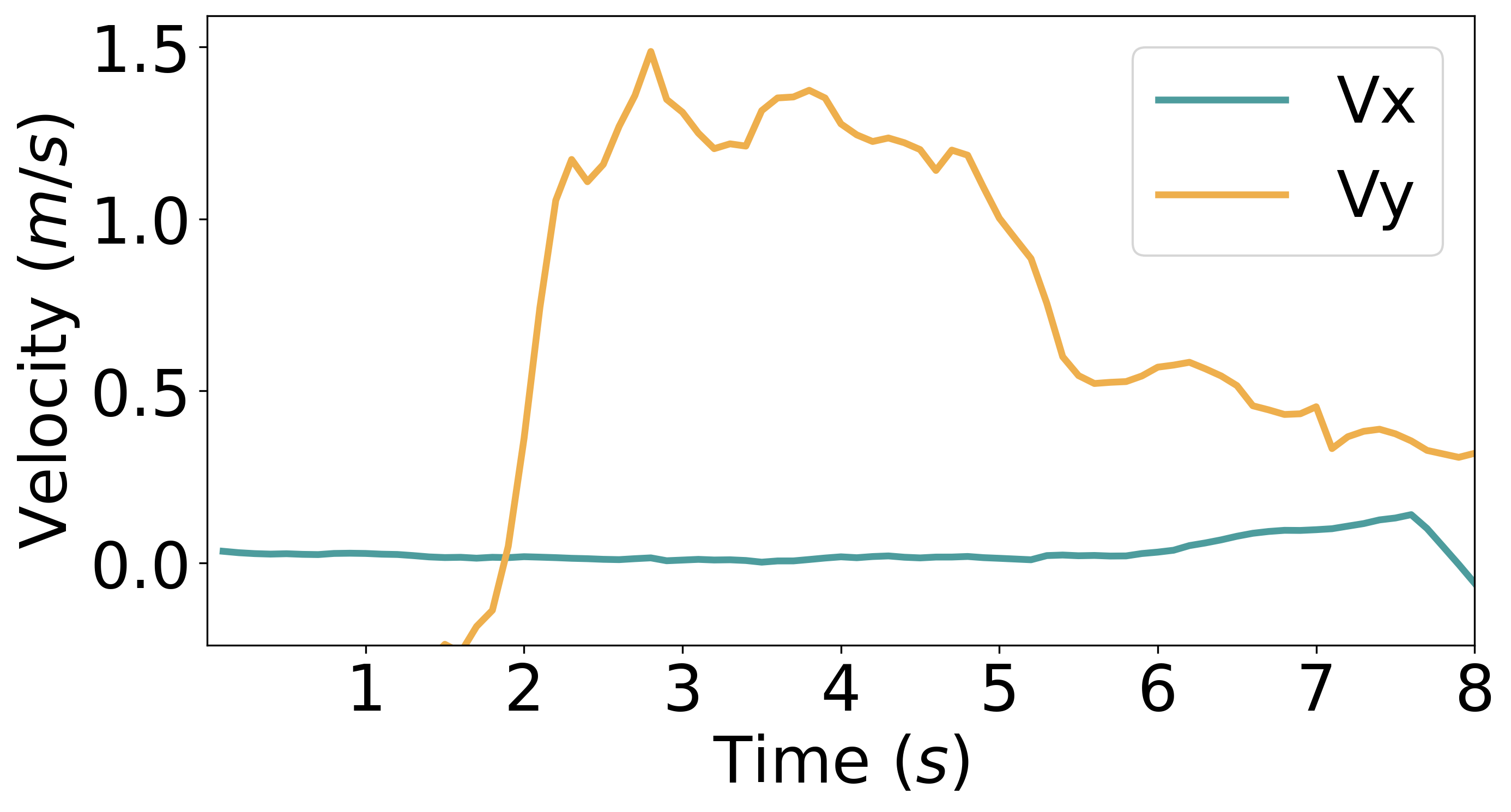}
    \end{subfigure}
    \begin{subfigure}{0.29\linewidth}
        \centering
        \includegraphics[width=\linewidth]{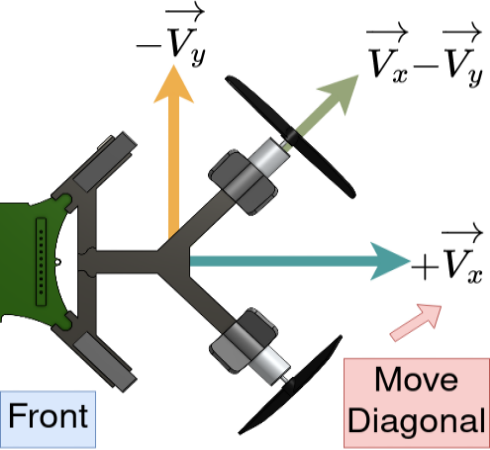}
    \end{subfigure}
    \begin{subfigure}{0.7\linewidth}
        \centering
        \includegraphics[width=\linewidth]{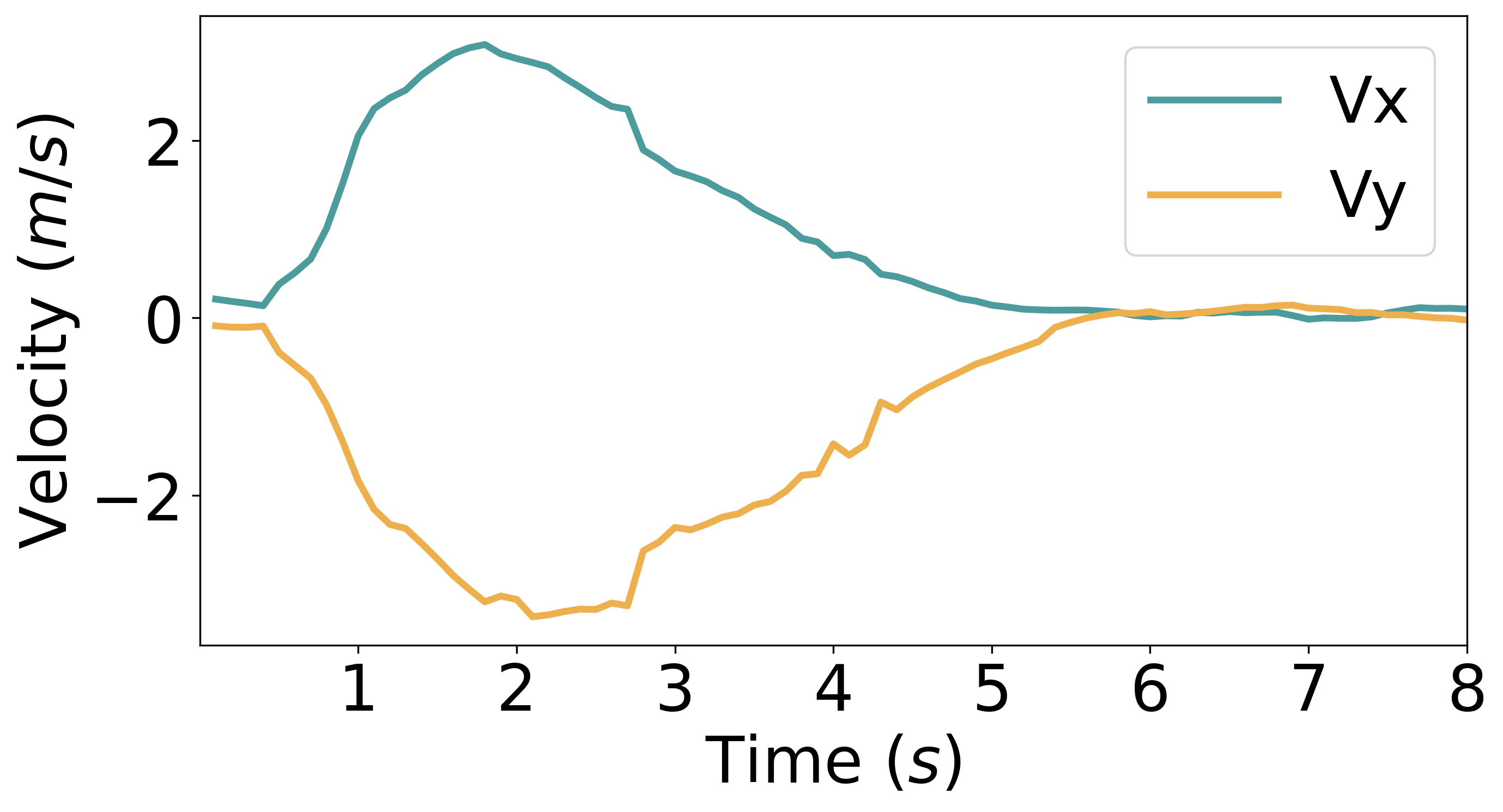}
    \end{subfigure}
    \vspace{-6pt}
    \caption{\beavis translation (horizontal) velocities during planar motion.}
    \label{fig:translation-speed}
\vspace{-12pt}
\end{figure}

We characterize the performance of the platform by using mobility and endurance metrics. 
\textcolor{black}{Setting a fair baseline for \beavis is nontrivial. While blimps are less maneuverable, the speed and angular velocity of quadrotors vary with their size. To carry the same payload as our platform (60g), a quadrotor will need to be significantly scaled in size and rotor power to even lift off.} In particular, we evaluate the translational velocity to get an idea of the planar motion and the rotational speed to indicate the turning capability. For endurance, we find the total flight time of \beavis which we compare against a traditional nano UAV. In our case, we use a crazyflie 2.1 which weighs 29.1\,g and has a similar form factor as our platform (without including the helium balloon). To keep the comparisons fair, both platforms use the same 300\,mAh Lipo battery.

\fakepar{Translation speed} We measure the speed with which \beavis can translate in a horizontal plane at a stable height. Fig.~\ref{fig:translation-speed} shows the typical speed attained by the platform for all the motion in the plane. We can see that, unlike traditional blimps, our platform is capable of motion in the \texttt{X-Y} plane depending on the desired direction. We achieve a velocity of 2.45\,m/s diagonally and 1.5\,m/s in the \texttt{X} and \texttt{Y} axes respectively. 
\textcolor{black}{In comparison, the horizontal speed varies between 0-2.1\,m/s for a Crazyflie 2.1, using similar rotors as our prototype, and between 0-8\,m/s for the Hummingbird drone, with similar dimensions as our prototype.}
\begin{figure}[!tb]
    \centering
    \includegraphics[width=0.85\linewidth]{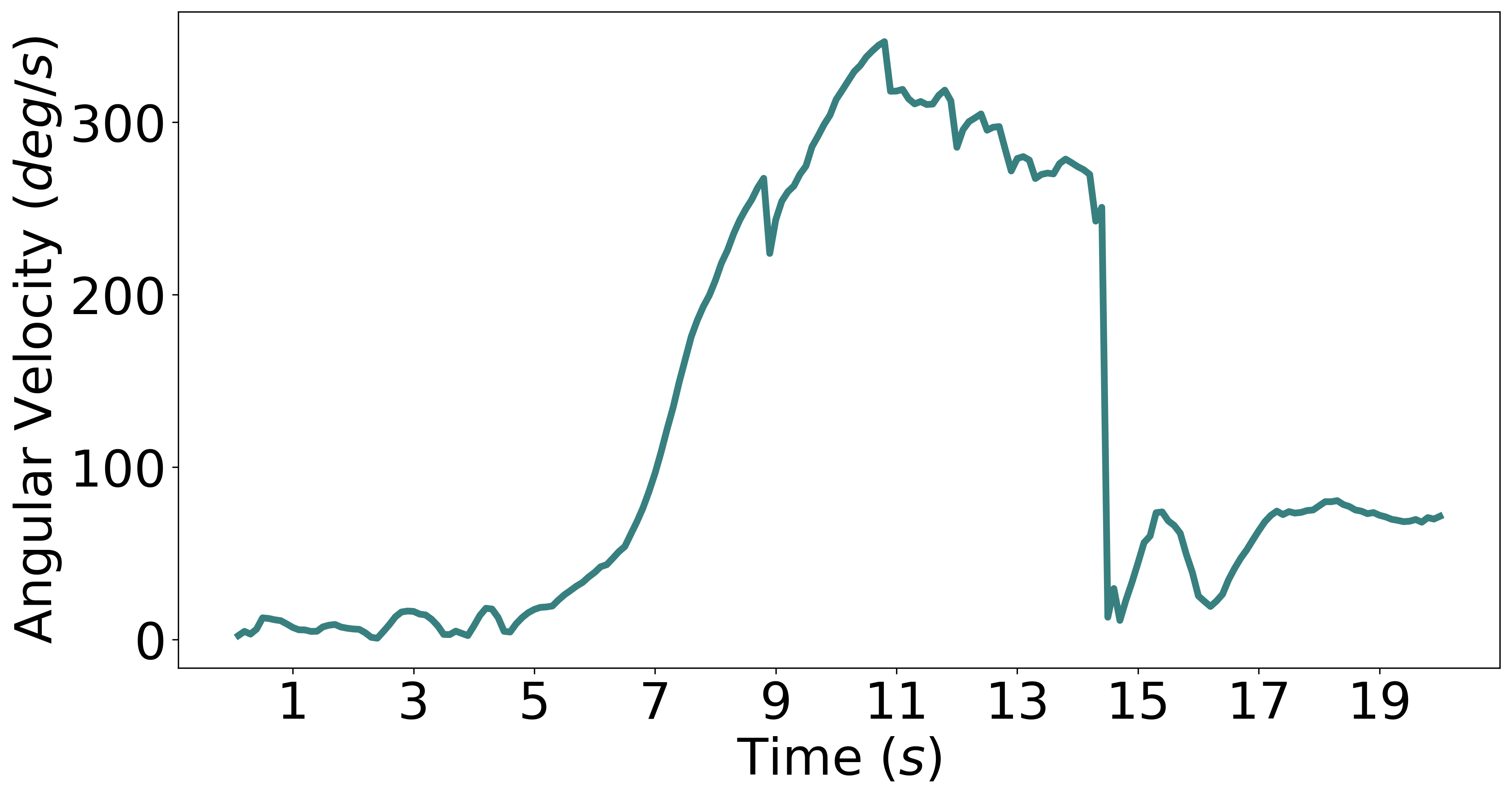}
    \caption{Rotational velocity (in yaw) of \beavis when yawing with maximum thrust.}
    \label{fig:rotational-speed}
\vspace{-10pt}
\end{figure}

\fakepar{Rotational speed and lifetime} We also measure the rotational velocity with which \beavis can turn in the yaw axis. Fig.~\ref{fig:rotational-speed} shows that a maximum value of \ang{346}\,/s rotational speed with maximum applied thrust velocity was attained.

\begin{figure}[!tb]
    \centering
    \includegraphics[width=0.9\linewidth]{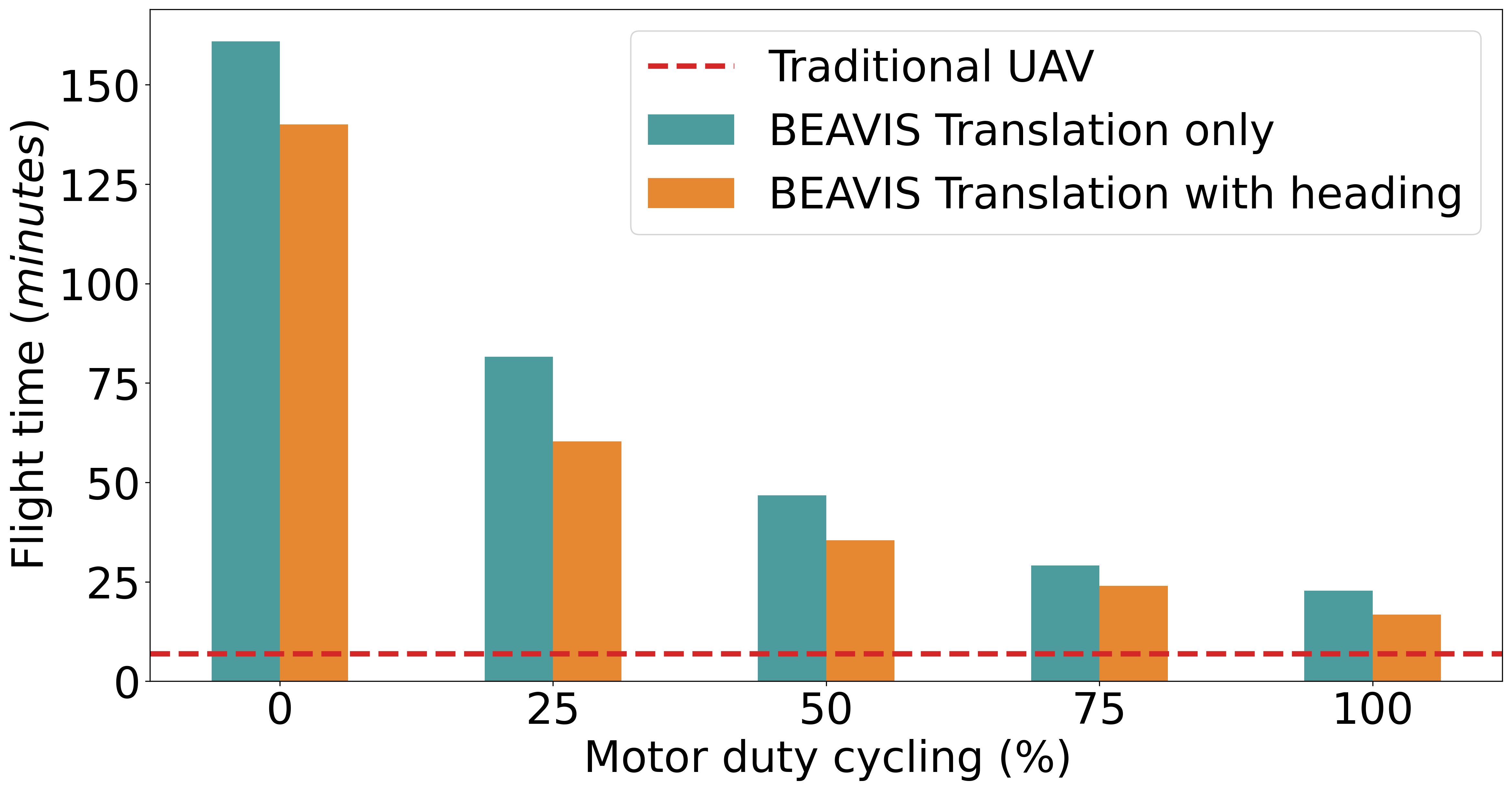}
    \vspace{-10pt}
    \caption{Flight time for \beavis for different flying methods and motor duty cycling.}
    \label{fig:battery-lifetime}
\vspace{-10pt}
\end{figure}

Lifetime is a pivotal metric of endurance. Fig.~\ref{fig:battery-lifetime} shows the maximum time a platform can operate before needing to land. We measure this time for two configurations, translation only, where the heading is not relevant and no yaw control is needed, and translation with a heading, where the platform actively maintains its orientation while moving. 

The graph shows the flight time with varying amounts of \textit{motor duty cycling}, which is only possible for our platform and not for any other UAVs. Even when operating at maximum thrust with 100\% duty cycle, the flight time is 16.7\,minutes while maintaining heading and 22.7\,minutes otherwise. This is an increase of at least 153\% (or 244\% without heading) from the 6.6\,minutes flight time of the similarly sized traditional UAV (denoted by a dashed red line). Further, with active duty cycling, we can increase this number up to 60.2\,minutes (812.1\% increase) while maintaining heading and to 81.6\,minutes (1136.3\% increase) otherwise. The case for hovering or 0\% duty cycle is also shown where the only limit to flight time comes from battery usage when communicating with the ground station (1\,Hz at 250\,kbps data rate) or additionally also for maintaining the attitude (heading case). Through our platform, we can extend the lifetime of a nano-sized UAV from a mere few minutes to over 2 hours. Lastly, we also carry about 60\,g of sample payload while flying which is impossible for the UAVs to do. This is equivalent to having 6 additional 300\,mAh LiPo batteries which can if required, enable a further 600\% increase in flight time.
\vspace{-7pt}
\subsection{Stability and Controllability}
\begin{figure}[!tb]
    \centering
    \begin{subfigure}{0.49\linewidth}
        \centering
        \includegraphics[width=\linewidth]{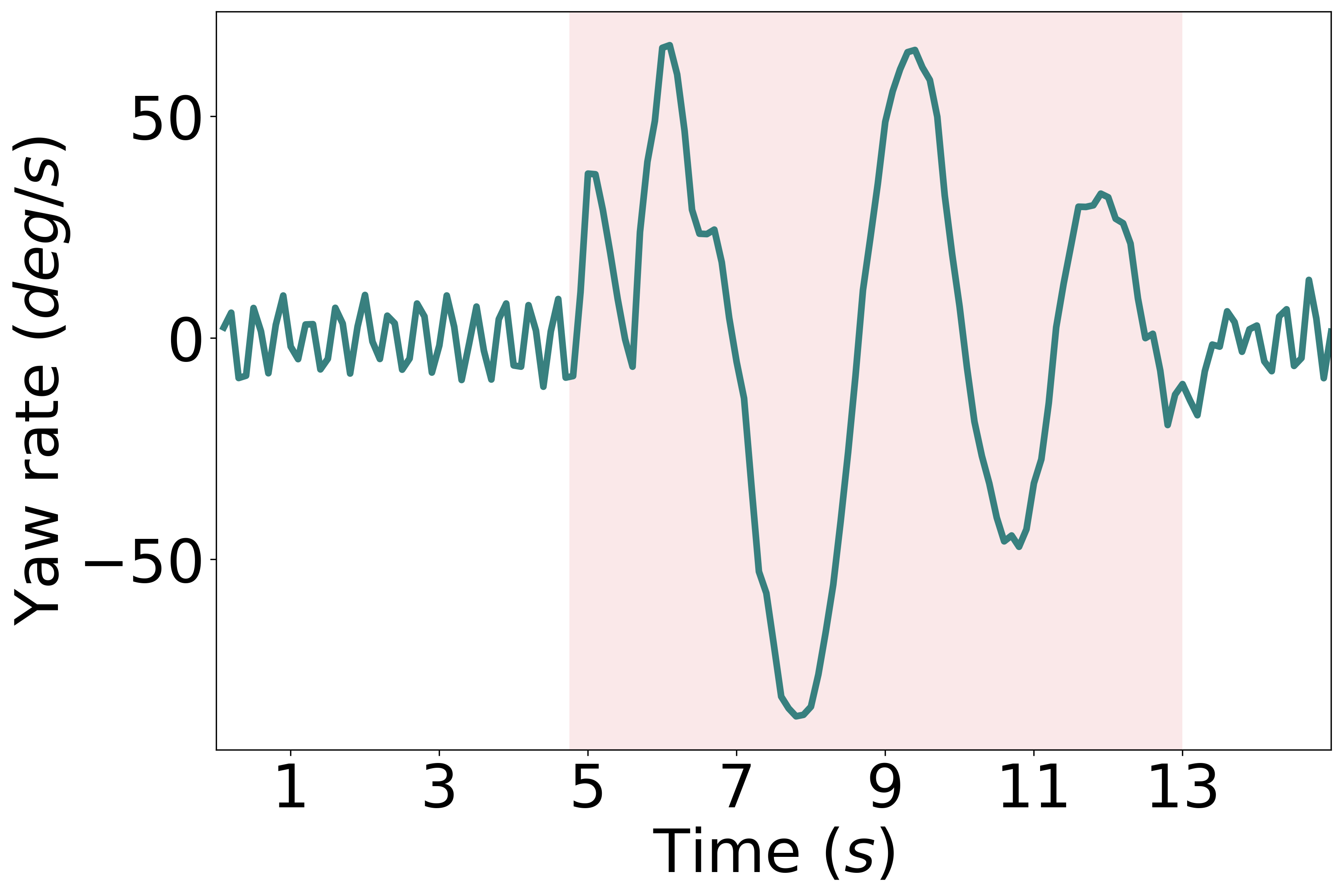}
    \end{subfigure}
    \begin{subfigure}{0.49\linewidth}
        \centering
        \includegraphics[width=\linewidth]{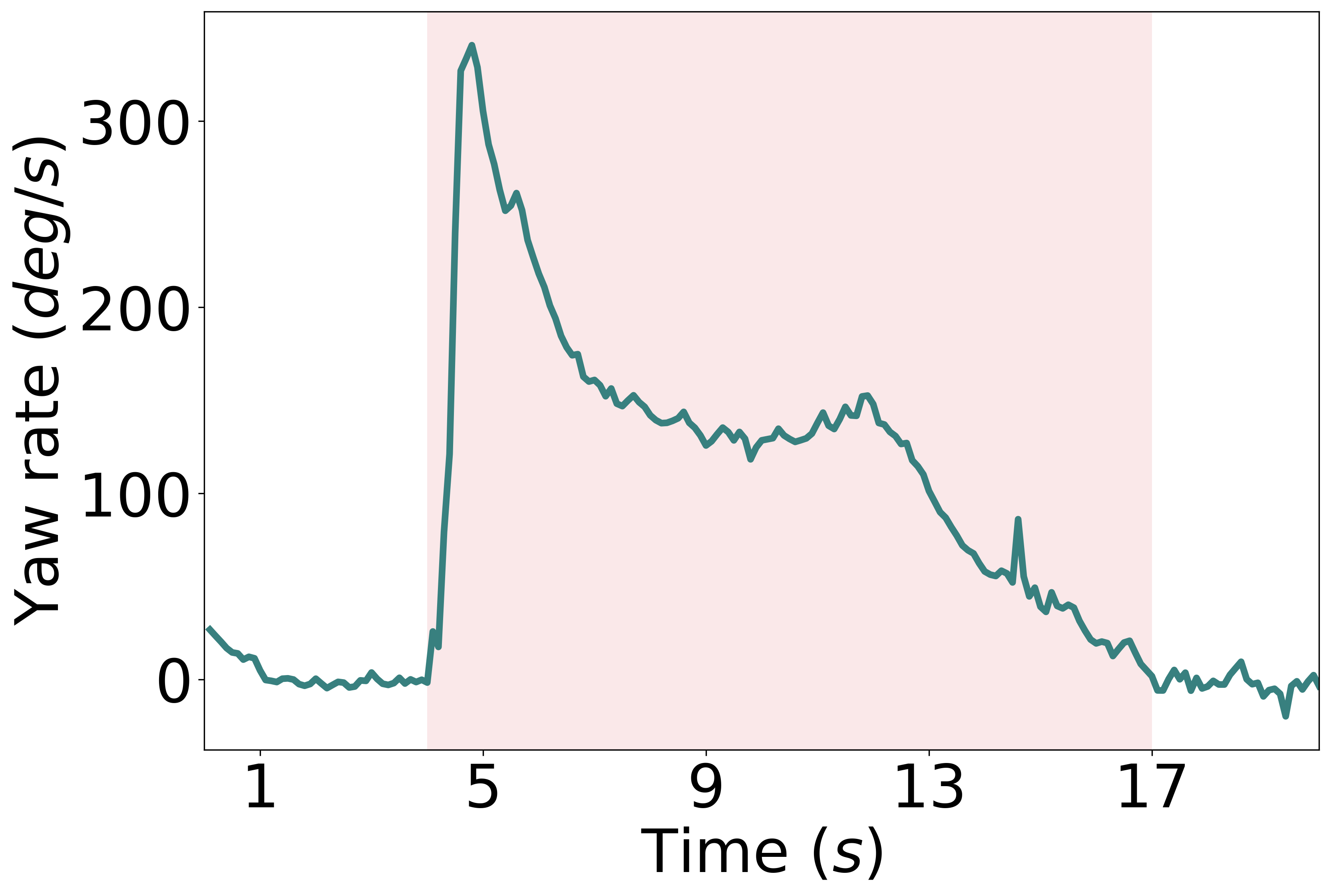}
    \end{subfigure}
    \begin{subfigure}{0.49\linewidth}
        \centering
        \includegraphics[width=\linewidth]{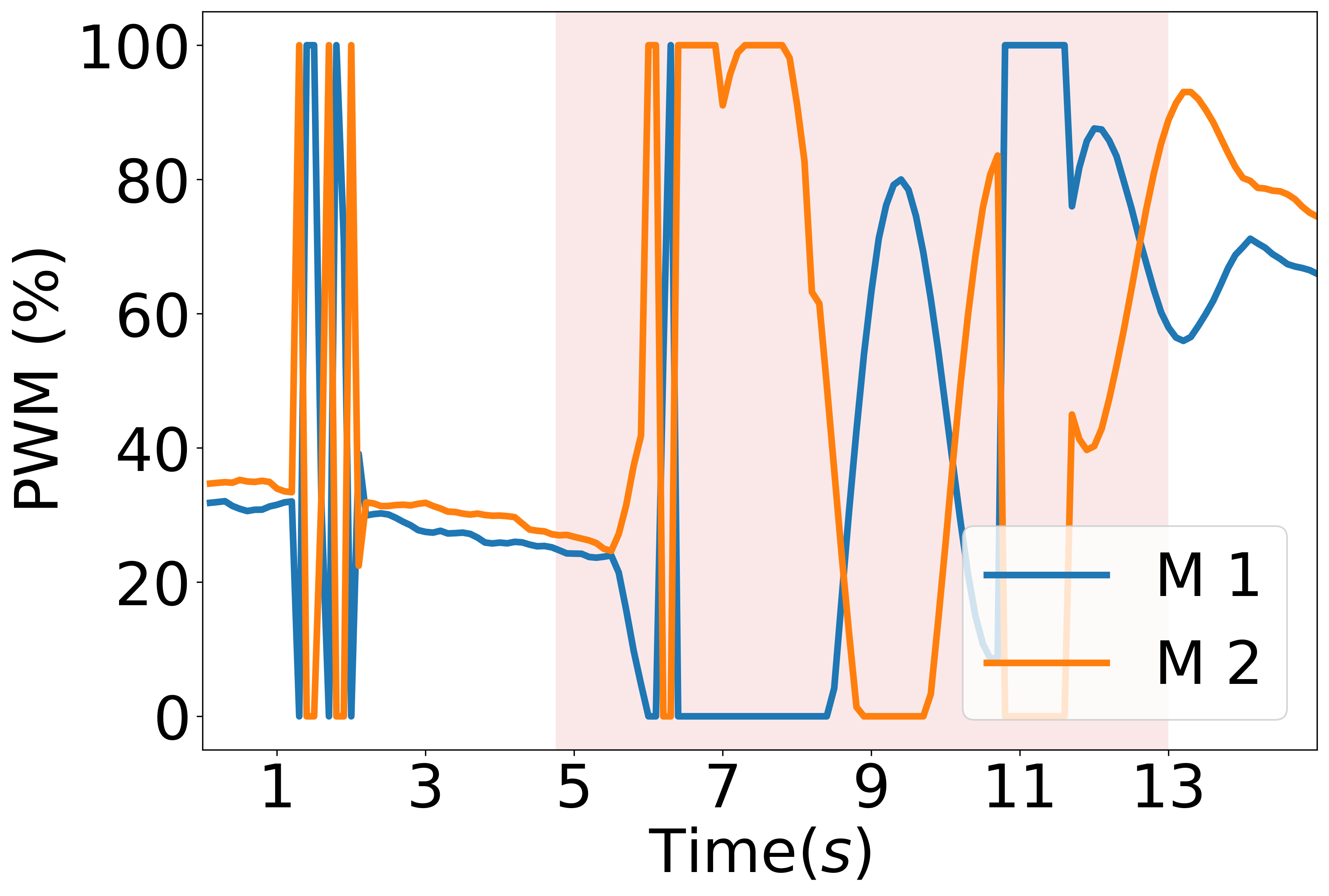}
    \end{subfigure}
    \begin{subfigure}{0.49\linewidth}
        \centering
        \includegraphics[width=\linewidth]{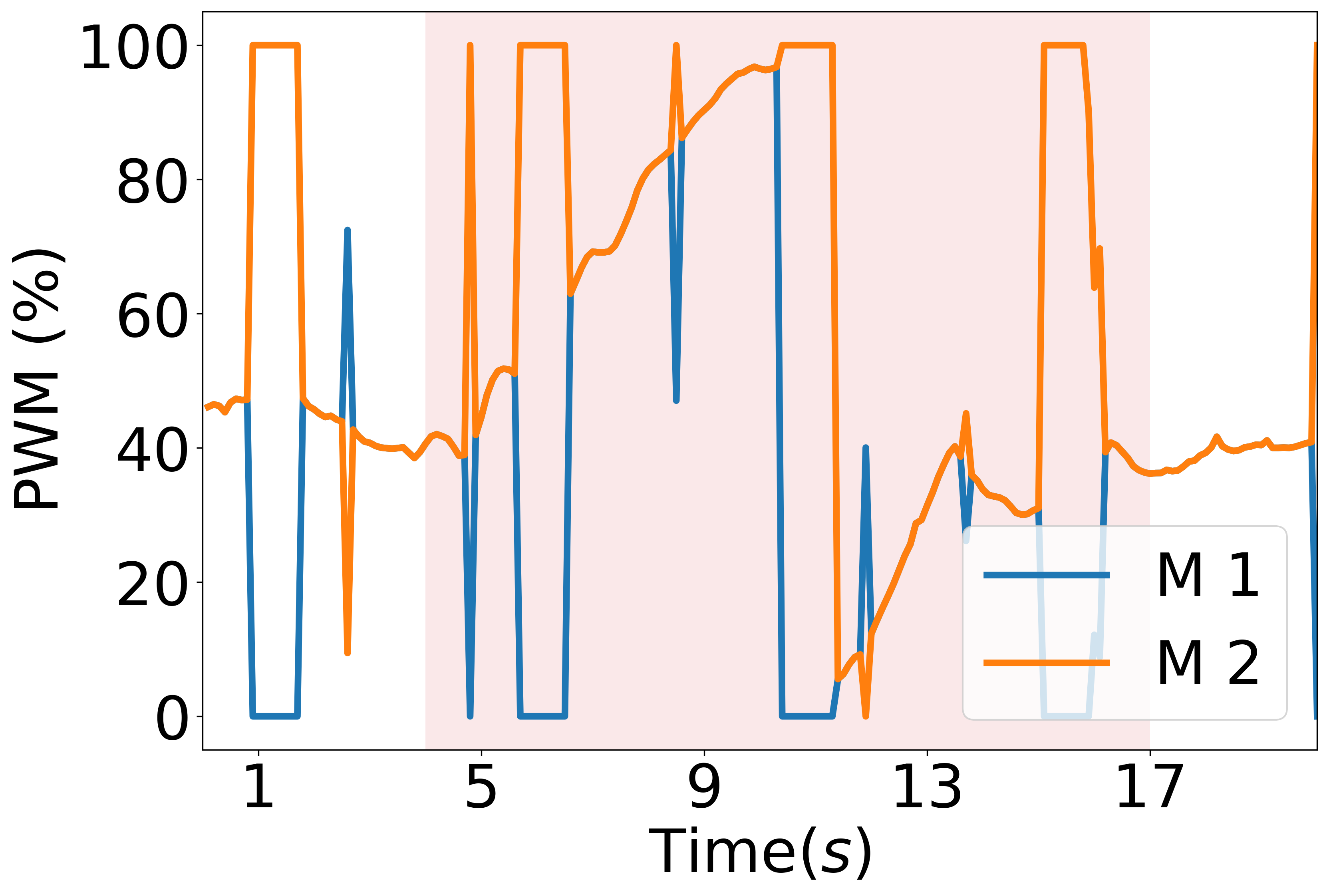}
    \end{subfigure}
    \caption{Stability of the platform when disturbances are introduced in the yaw axis while hovering. The two rates of induced spin are shown: low (left) and high (right) along with the motor signal.}
    \label{fig:yaw-stab}
\vspace{-12pt}
\end{figure}
We evaluate the stability and performance of the controller. The design of the platform offers a natural stability in the pitch and roll axes and also contributes to the dynamics. As mentioned before, the motion dynamics of this platform are highly nonlinear, and thus, it is interesting to note how well the control algorithm can fly this platform. We also evaluate the performance of our fault-sensing algorithm.

\fakepar{Yaw stability against vortexes} We show the response of \beavis when instability is introduced (in the form of spin) in the yaw (heading) axis. This spin could naturally be induced by wind gusts or vortexes. Fig.~\ref{fig:yaw-stab} shows how the platform stabilizes after the initial large angular velocity caused by the spin (in red). The speed and magnitude of the response to this instability can be set by tuning the parameters of the controller based on the necessity of applications.
\begin{figure}[!tb]
    \centering
    \begin{subfigure}{0.49\linewidth}
        \centering
        \includegraphics[width=\linewidth]{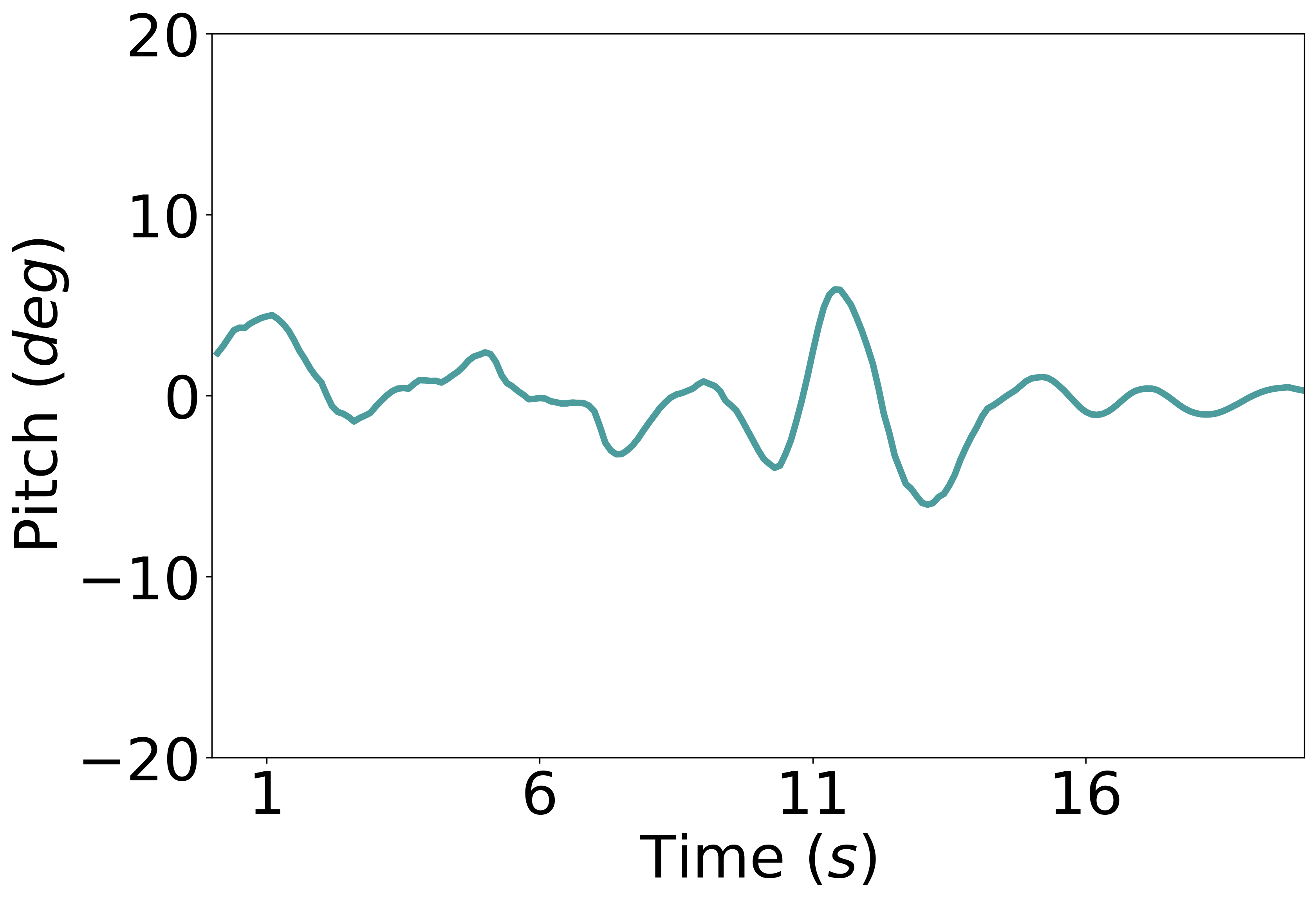}
    \end{subfigure}
    \begin{subfigure}{0.49\linewidth}
        \centering
        \includegraphics[width=\linewidth]{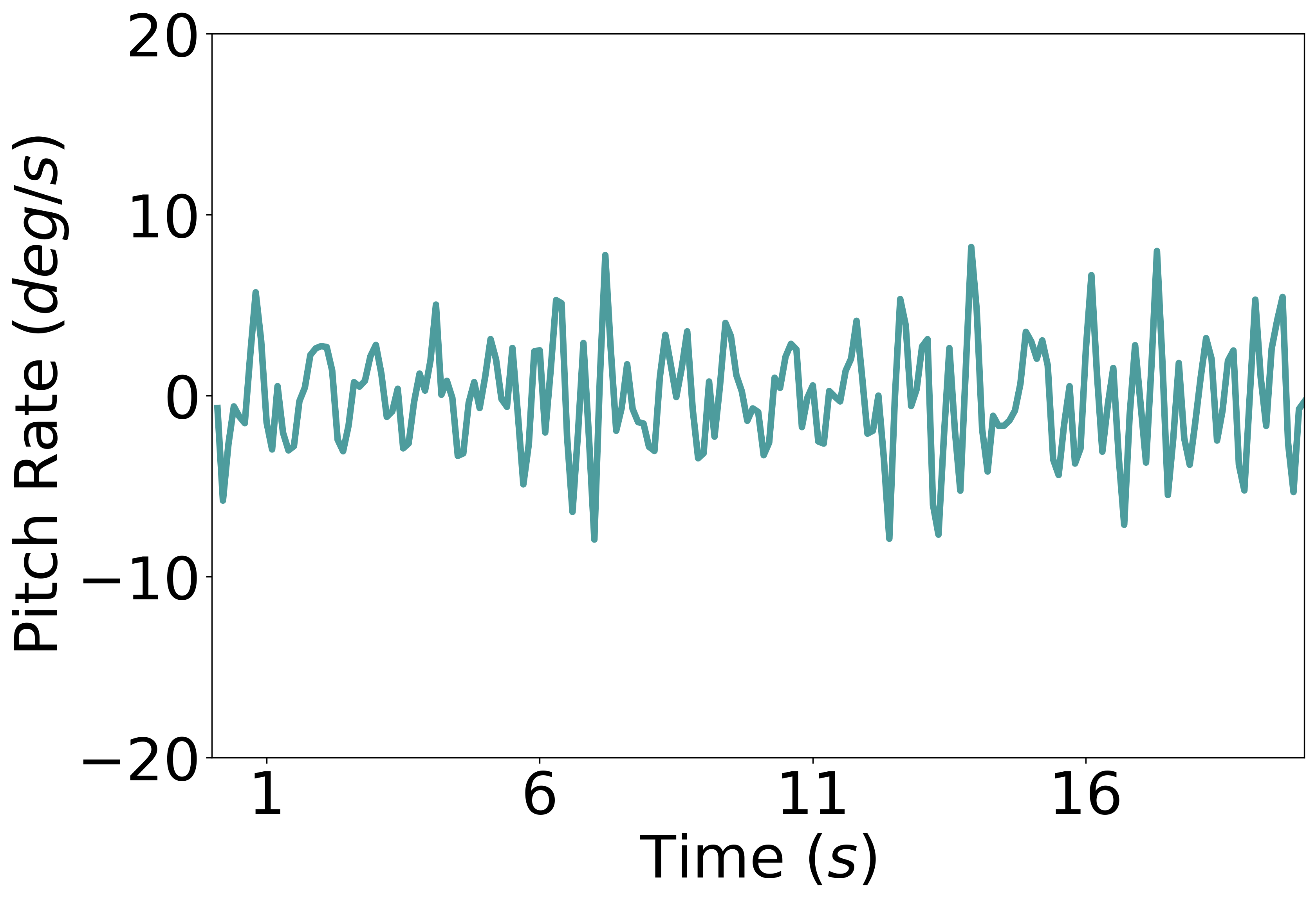}
    \end{subfigure}
    \begin{subfigure}{0.49\linewidth}
        \centering
        \includegraphics[width=\linewidth]{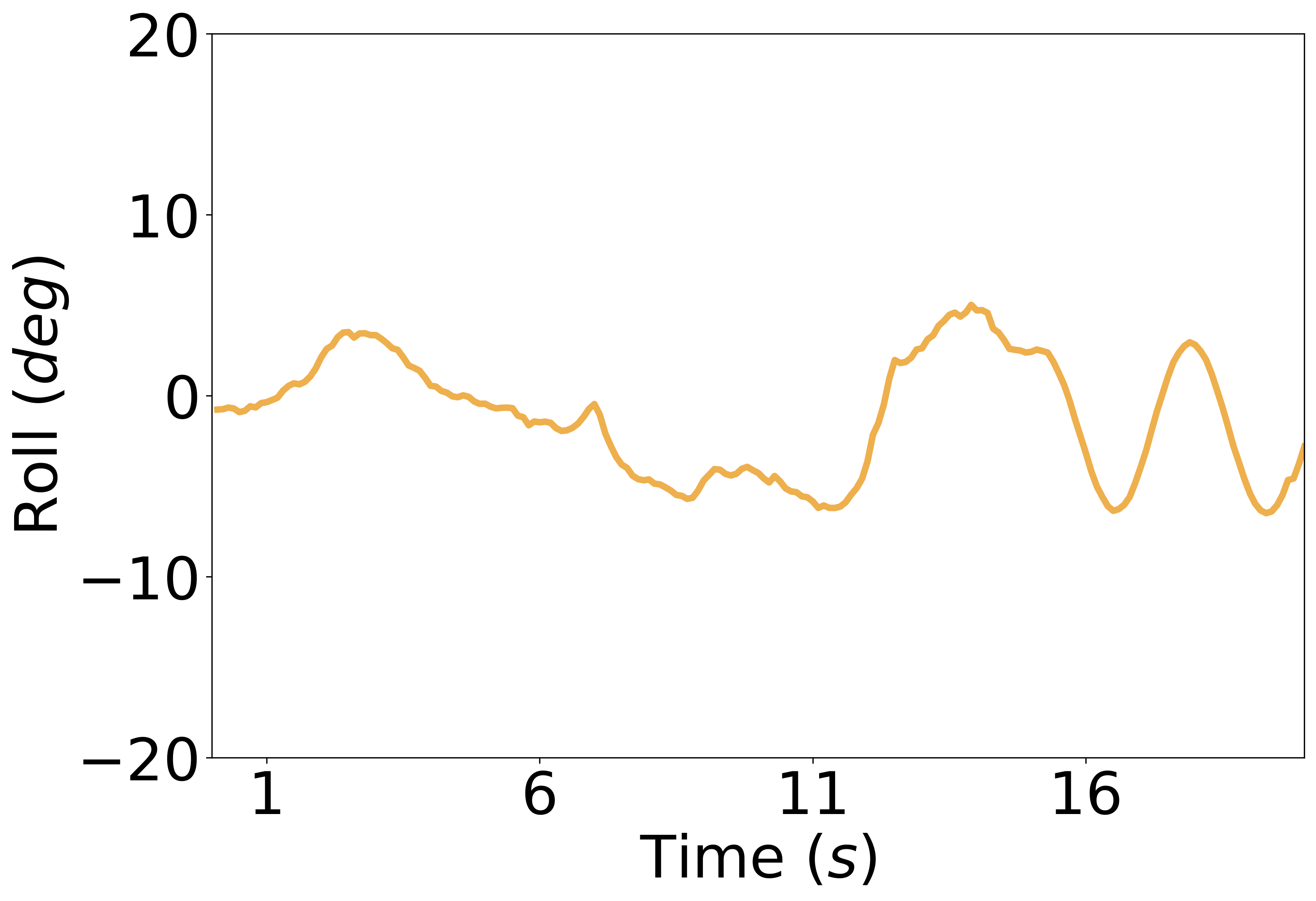}
    \end{subfigure}
    \begin{subfigure}{0.49\linewidth}
        \centering
        \includegraphics[width=\linewidth]{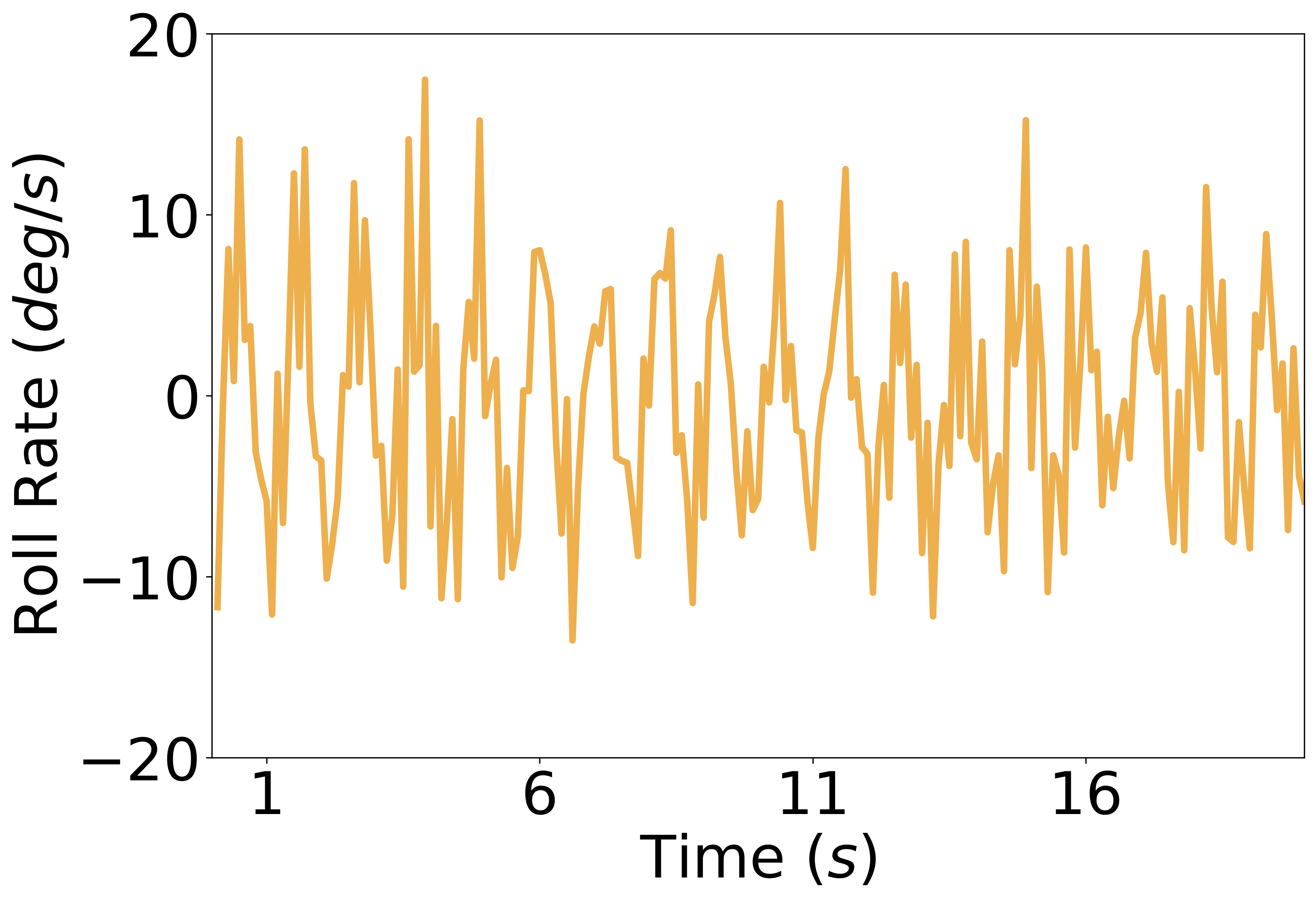}
    \end{subfigure}
    \caption{Stability of the platform in the pitch (\texttt{X}) and roll (\texttt{Y}) angles and angular rates when moving in a straight line for 20\,s.}
    \label{fig:pitch-roll-stab}
\vspace{-5pt}
\end{figure}

\fakepar{Roll and pitch stability} We show the angular stability of the platform in the horizontal plane, (\texttt{X, Y}), corresponding to pitch and roll angles when translating. Fig.~\ref{fig:pitch-roll-stab} shows the variation of pitch angle, pitch angular velocity, roll angle, and roll angular velocity when moving in a straight line for 20\,s. The drag introduced by the balloon ensures that the platform remains naturally stable in these axes.
\begin{figure}[!tb]
    \centering
    \includegraphics[height=0.35\linewidth, width=0.85\linewidth]{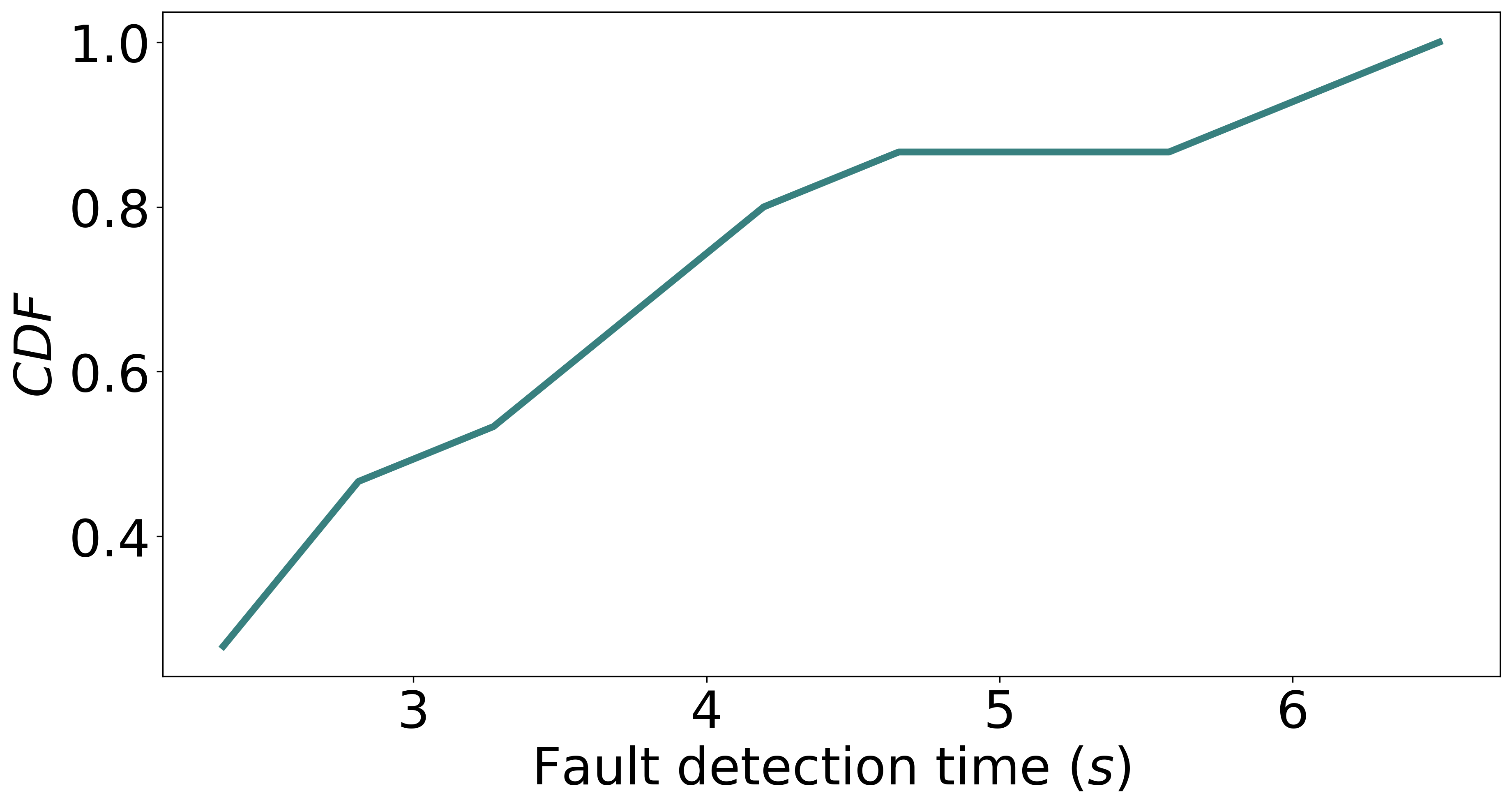}
    \caption{CDF of rotor fault detection time after a rotor fault is induced mid-flight.}
    \label{fig:FTC-sensing}
\vspace{-10pt}
\end{figure}

\fakepar{Rotor fault sensing} We evaluate the time delay with which our algorithm can detect rotor failures. In these experiments, we start with a regular platform with all 4 motors and propellers working normally. Once an equilibrium, stable state is achieved, we send a wireless command to disable a single motor. Fig.~\ref{fig:FTC-sensing} shows that our algorithm can detect motor failures within 5.5\,s in 90\% of the cases.
\begin{figure}[!tb]
    \centering
    \begin{subfigure}{0.37\linewidth}
        \centering
        \includegraphics[height=\linewidth,width=\linewidth]{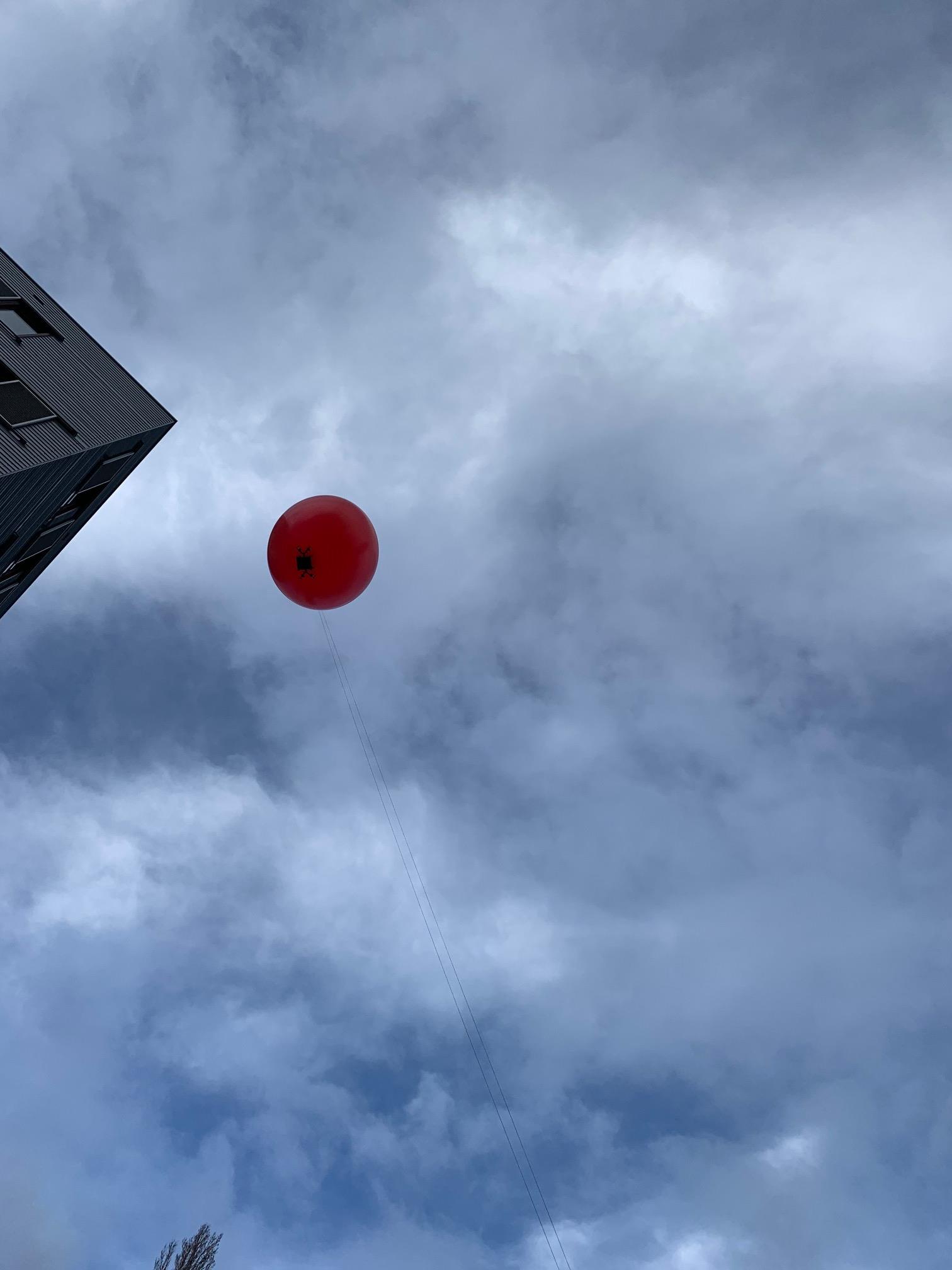}
    \end{subfigure}
    \begin{subfigure}{0.37\linewidth}
        \centering
        \includegraphics[height=\linewidth,width=\linewidth]{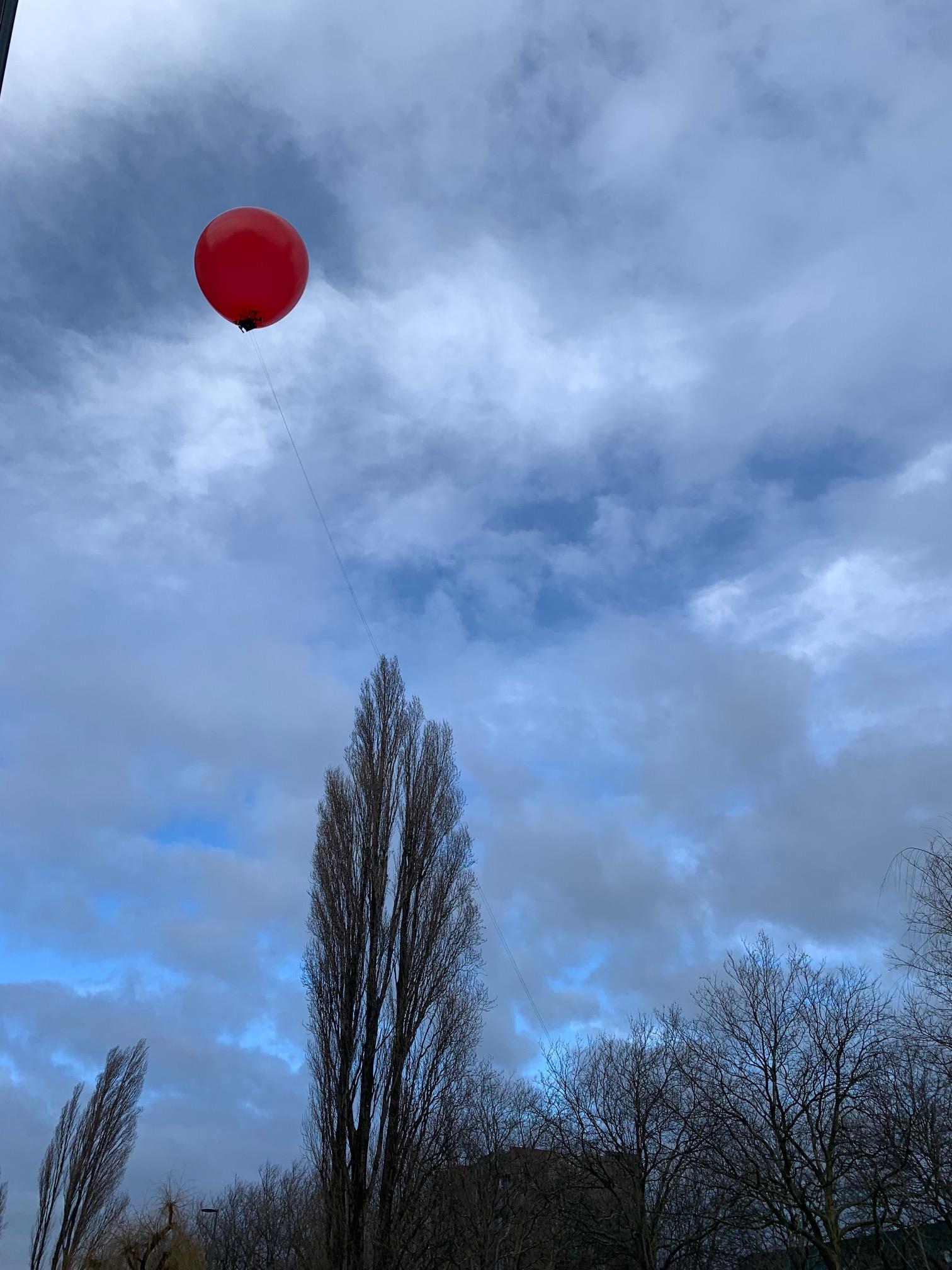}
    \end{subfigure}
    \caption{Tethered \beavis flights outdoors in the presence of varying wind speeds.}
    \label{fig:outdoor-tether}
\vspace{-8pt}
\end{figure}
\vspace{-5pt}
\subsection{Outdoor Operation and Wind}
\textcolor{black}{When flying outdoors, weather phenomena like strong wind gusts and crosswinds might need to be endured. Wind gusts in particular can induce rotational velocity that causes the balloon to spin and wobble. The angular stability of \beavis for pitch and roll disturbances when flying is within $10^\circ$. We observed that the buoyancy of the balloon added resilience to mild crosswinds, up to 3\,m/s, and it did not impede vertical motion. Upon exposure to strong continuous wind, exceeding 6\,m/s, \beavis loses its heading and needs to compensate by flying back to its target hover location, similar to other blimps.}
\beavis is capable of adapting to wind disturbances when flown in similar conditions using our propeller design. This response is quantified as a key performance metric like angular recovery (8s for 50\,degrees/s angular rate). Based on the volume of the vehicle, the maximum thrust of a single rotor and the drag of the balloon, and the limits of maximum external wind speed with the current \beavis prototype, we can fly against 8 - 8.5\,m/s, as governed by the current thrust-drag ratio. At these speeds, the prototype is unable to recover its target position and keeps getting blown away.

\textcolor{black}{Owing to its design however, BEAVIS has higher maneuverability (extra degrees of freedom), direct actuation in directions not possible before (sideways, vertical), and higher agility (making quicker, sharper motions). Even though traditional blimps suffer from non-holonomic constraints and restricted degrees of movement, their operating conditions are well-defined. This translates for our platform as well, up to 100\,m altitude, \beavis can be considered analogous to airships; and will be able to fly at their operating temperature, pressure, and humidity profiles. The learning-based algorithm in BEAVIS can adapt to variable flight dynamics caused by weather variations (temperature, humidity, pressure) using a dynamic control loop.}

\textcolor{black}{We carry out tethered flight tests, using a string, to verify this behavior as seen in Fig.~\ref{fig:outdoor-tether}. A tether interferes with flight dynamics significantly as we fly close to neutral buoyancy (zero weight) introducing a pulling-down effect. Even though performance is suboptimal, we include the tethered tests performed outside. The use of a tether to conduct outdoor tests is because of our testing location and the limitations imposed by government authorities. Tethered deployment of \beavis may be useful in some applications such as communication support in disaster zones. Note the current model uses small coreless motors with only 15\,g of maximum thrust. This limits its speed and ability to withstand large wind gusts. Note: The western region of the Netherlands experiences regular high ambient wind speeds resulting in the current prototype flying away when subjected to windy conditions. This makes the uncontrolled outdoor environment very challenging for experimentation.}
\vspace{-15pt}
\section{Discussions}
\label{sec:discussion}
\fakepar{Further applicability} 
\textcolor{black}{BEAVIS can track people both indoors and outdoors since it can be scaled with the required rotor power. The aerodynamic effects are the same as in our CFD simulations. Our prototype used a balloon with a 55\,cm diameter, which can be reduced by optimizing the payload and/or by miniaturization, making it suitable for indoors. The size and maneuverability of the platform enable indoor applications, for instance, continuously localizing gas leaks in factories by allowing navigation through tight spaces. While the size of blimps is typically 1\,m, our platform imposes no such limitations on the size.} 

In outdoor settings, our design may be used for weather prediction and monitoring. It is very easy to add humidity, temperature, and wind speed sensors to \beavis while it can use our control algorithm to stabilize itself under varying environmental conditions. \beavis is optimized for hovering, applications like wildlife tracking or poacher detection. The platform can follow animals from a high altitude without disturbing the creatures because of its passive hovering. It can also be used for wildfire detection, where it can stay hovering for long periods (helium being inert does not cause a fire). There is a growing trend to deploy more sensors in the sky to enable this after recent wildfires wreaked havoc around the world~\cite{wildfire_2022}. Lastly, for disaster scenarios, like volcanic eruptions or earthquakes, multiple \beavis platforms can be commissioned together to create ad hoc networks for rescue and connectivity. Such calamities can cause a blackout of communication for days~\cite{tonga-2022}. 

\fakepar{Other salient features} 
The \beavis platform is designed to operate quietly and safely. It doesn't make a loud noise when hovering at a spot unless constant stabilization is needed due to wind flow. The balloon adds sufficient drag so that the platform doesn't accelerate to dangerous speeds while safeguarding against temporal wind gusts. It's also safe to operate around people because of the relatively small size of the motors and propellers used. The design of \beavis enables fault-tolerant operation inherently as loss of a single or more rotor does not catastrophically inhibit our platform as it would in the case of traditional UAVs. \beavis is still controllable compared to a blimp when rotors fail. It's also relatively easy to add additional payload without compromising on performance. The design of the platform uses relatively simple components, making it accessible.

\fakepar{Limitations} 
One of the consequences of enhancing the endurance by using a helium-filled balloon is the significantly larger footprint of the platform which limits the places where \beavis can reach when compared to other nano-UAVs. Further, the top speed of the platform is limited due to the drag introduced by the balloon. 
\textcolor{black}{The additional drag caused by the balloon is a known effect in the trade-off between endurance and speed.} It is possible to change the speed by using more powerful motors but that comes at the cost of a reduced lifetime and possibly a larger balloon size to account for the additional weight. Commercial latex balloons are imperfect and the filled helium escapes over time (in a matter of 48$\,hours$). This means that the platform cannot hover for too long. While it is possible to improve the balloon design and explore more efficient methods of filling them or using foil-based alternatives, that remains outside the scope of this work. We sourced only easily available components to make \beavis generic and easy to replicate. Further, the platform is also not very stable against extreme wind gales which can easily blow the balloon off of its set location. It is possible to bring the balloon back but at the cost of increased energy usage. Despite these limitations, we believe \beavis holds great potential for the sensing and IoT community.
\vspace{-15pt}
\section{Conclusion}
\label{sec:end}
We developed an aerial platform called \beavis that can be used in different sensing applications. We provided a detailed explanation and characterization of the new aerodynamic flow phenomenon that \beavis uses for gaining all degrees of freedom akin to a drone despite being under-actuated. We discussed the design methodology and system components to build a \beavis prototype. Using computational fluid dynamics calculations it is easy to generalize the same phenomenon for different sizes of the platform. To enable autonomous flight, we also designed a new data-driven flight control algorithm that can detect rotor faults and adjust the control logic to safely land. We built a prototype using COTS hardware to evaluate the performance in two applications showcasing all the capabilities of \beavis. 

It achieves a translation and rotational speed of 2.45\,m/s in the X-Y plane and \ang{346}/s, respectively. It also achieves a lifetime increase of 1136.3\% compared to identical UAVs with complete planar and vertical motion with yaw control. Our algorithm could detect a rotor fault within 5.5\,s in 90\% of the cases. 
The next step is to make \beavis more robust and find the trade-off between helium required for lifting a particular weight, motor power to negotiate wind speeds, size, and response time.

\bibliographystyle{ACM-Reference-Format}
\bibliography{references}

\end{document}